\documentclass[11pt]{article}

\usepackage[preprint]{acl}

\usepackage{times}
\usepackage{amsfonts}
\usepackage{latexsym}
\usepackage[T1]{fontenc}

\usepackage{stfloats}
\usepackage[utf8]{inputenc}
\usepackage{array} 
\usepackage[most]{tcolorbox}
\usepackage{cleveref}
\usepackage[table]{xcolor}
\usepackage{amsmath}
\usepackage{siunitx}
\usepackage{multirow}
\usepackage{microtype}
\usepackage{booktabs,tabularx,ragged2e,array}
\usepackage{inconsolata}
\usepackage{subcaption}
\usepackage{graphicx}
\usepackage{pifont}
\usepackage{makecell}

\newcommand{\dplus}[1]{\,{\scriptsize\textcolor{red}{(+#1)}}}
\newcommand{\dminus}[1]{\,{\scriptsize\textcolor{blue}{(-#1)}}}
\newcommand{\GroupHeader}[1]{%
  \rowcolor{black!5}\multicolumn{6}{l}{\textbf{\textit{#1}}}\\[-2pt]
  \cmidrule{1-6}
}
\DeclareUnicodeCharacter{266B}{}
\title{MARCH: Evaluating the Intersection of Ambiguity Interpretation and Multi-hop Inference}

\author{
\textbf{Jeonghyun Park}$^{1} \footnotemark[1]$ , 
\textbf{Ingeol Baek}$^{1} \thanks{Equal contribution. Authors randomized.}$,
\textbf{Seunghyun Yoon}$^{2}$,
\textbf{Haeun Jang}$^{1}$,
\textbf{Aparna Garimella}$^{3}$, \\
\textbf{Akriti Jain}$^{3}$,
\textbf{Nedim Lipka}$^{2}$,
\textbf{Hwanhee Lee}$^{1}\thanks{Corresponding author}$
\\
$^{1}$Chung-Ang University, Seoul, Korea, $^{2}$Adobe Research, USA, $^{3}$Adobe Research, India \\
\texttt{\{tom0365, ingeolbaek, jhe020814, hwanheelee\}@cau.ac.kr} \\
\texttt{\{syoon, garimell, akritij, lipka\}@adobe.com} \\
\url{https://jeonghyunpark2002.github.io/MARCH_project_page}
}

\begin{document}
\maketitle

\begin{abstract}
Real-world multi-hop QA is naturally linked with ambiguity, where a single query can trigger multiple reasoning paths that require independent resolution. Since ambiguity can occur at any stage, models must navigate layered uncertainty throughout the entire reasoning chain. 
Despite its prevalence in real-world user queries, previous benchmarks have primarily focused on single-hop ambiguity, leaving the complex interaction between multi-step inference and layered ambiguity underexplored.
 In this paper, we introduce \textbf{MARCH}, a benchmark for their intersection, with 2,209 multi-hop ambiguous questions curated via multi-LLM verification and validated by human annotation with strong agreement.
Our experiments reveal that even state-of-the-art models struggle with MARCH, confirming that combining ambiguity resolution with multi-step reasoning is a significant challenge. To address this, we propose \textbf{CLARION}, a two-stage agentic framework that explicitly decouples ambiguity planning from evidence-driven reasoning, significantly outperforms existing approaches, and paves the way for robust reasoning systems. The Code is available at \url{https://github.com/jeonghyunpark2002/MARCH.git}

\end{abstract}

\section{Introduction}

\newcommand{\cellcp}[2]{%
  \begin{tabular}{@{}c@{}}%
    \textbf{#1}\\[-2pt]%
    {\scriptsize(#2)}%
  \end{tabular}%
}

\begin{figure}[!ht]
    \centering
    \includegraphics[width=\columnwidth]{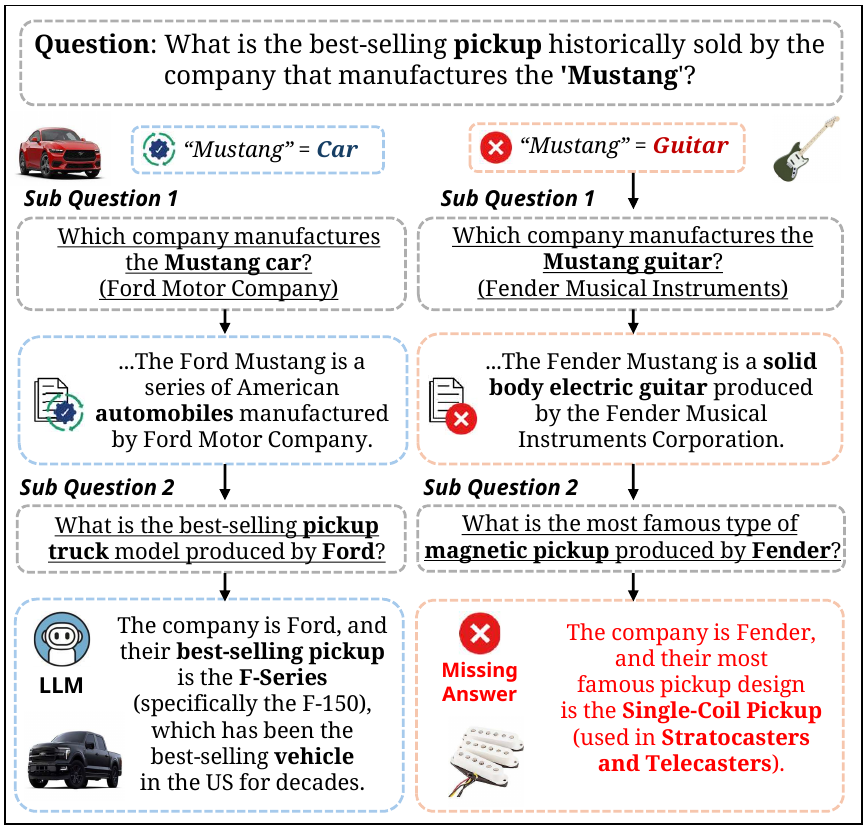}
    \vspace{-6mm}
    \caption{An example of multi-hop ambiguity QA. The ambiguity of the second hop ("pickup") is latent; it is only detectable if the alternative interpretation of the first hop ("Mustang" as guitar) is preserved.}
    \vspace{-7mm}
    \label{fig:challenge2}
\end{figure}

Multi-hop Question Answering (QA) presents a significant reasoning challenge, requiring models to construct logical chains by connecting disparate pieces of information scattered across multiple documents~\citep{trivedi2022musique, yang2018hotpotqa, ho2020constructing}. Ambiguity in QA can further complicate this process, as a single query may stem from polysemous terms or insufficient context, demanding clarification or interpretation before an answer can be derived~\citep{min-etal-2020-ambigqa}. The intersection of these two challenges---multi-hop reasoning and ambiguity---creates a uniquely difficult setting where uncertainty scales exponentially. In multi-hop ambiguous QA, ambiguity can emerge at any step of the reasoning chain, often remaining latent until prior steps are resolved. This interdependence means that errors in resolving early-stage ambiguity propagate downstream, causing models to prematurely commit to incorrect reasoning paths and producing incomplete or flawed answers.

\begin{figure}[t]
\captionsetup{skip=2pt}      
\setlength{\textfloatsep}{6pt}
\setlength{\floatsep}{4pt}
\setlength{\intextsep}{6pt}

  \centering
  \vspace{-4pt} 

  \begin{subfigure}[t]{0.9\columnwidth}
    \centering
    \footnotesize
    \setlength{\tabcolsep}{4pt}      
    \renewcommand{\arraystretch}{1.08} 

    \begin{tabular}{c c c c}
      \toprule
      & \textbf{Single-hop} & \textbf{Multi-hop} & \textbf{Total} \\
      \midrule
      \textbf{Ambiguous} &
        \cellcolor{orange!25}\cellcp{1010}{48.4\%} &
        \cellcolor{purple!20}\cellcp{277}{13.3\%} &
        \cellcp{1287}{61.7\%} \\
      \textbf{Non-ambig} &
        \cellcolor{gray!20}\cellcp{428}{20.5\%} &
        \cellcolor{blue!15}\cellcp{370}{17.7\%} &
        \cellcp{798}{38.3\%} \\
      \midrule
      \textbf{Total} &
        \cellcp{1438}{69.0\%} &
        \cellcp{647}{31.0\%} &
        \cellcp{2085}{100\%} \\
      \bottomrule
    \end{tabular}
  \end{subfigure}%


  \begin{subfigure}[t]{0.96\columnwidth}
    \centering
    \includegraphics[width=\columnwidth,trim=0 6 0 6,clip]{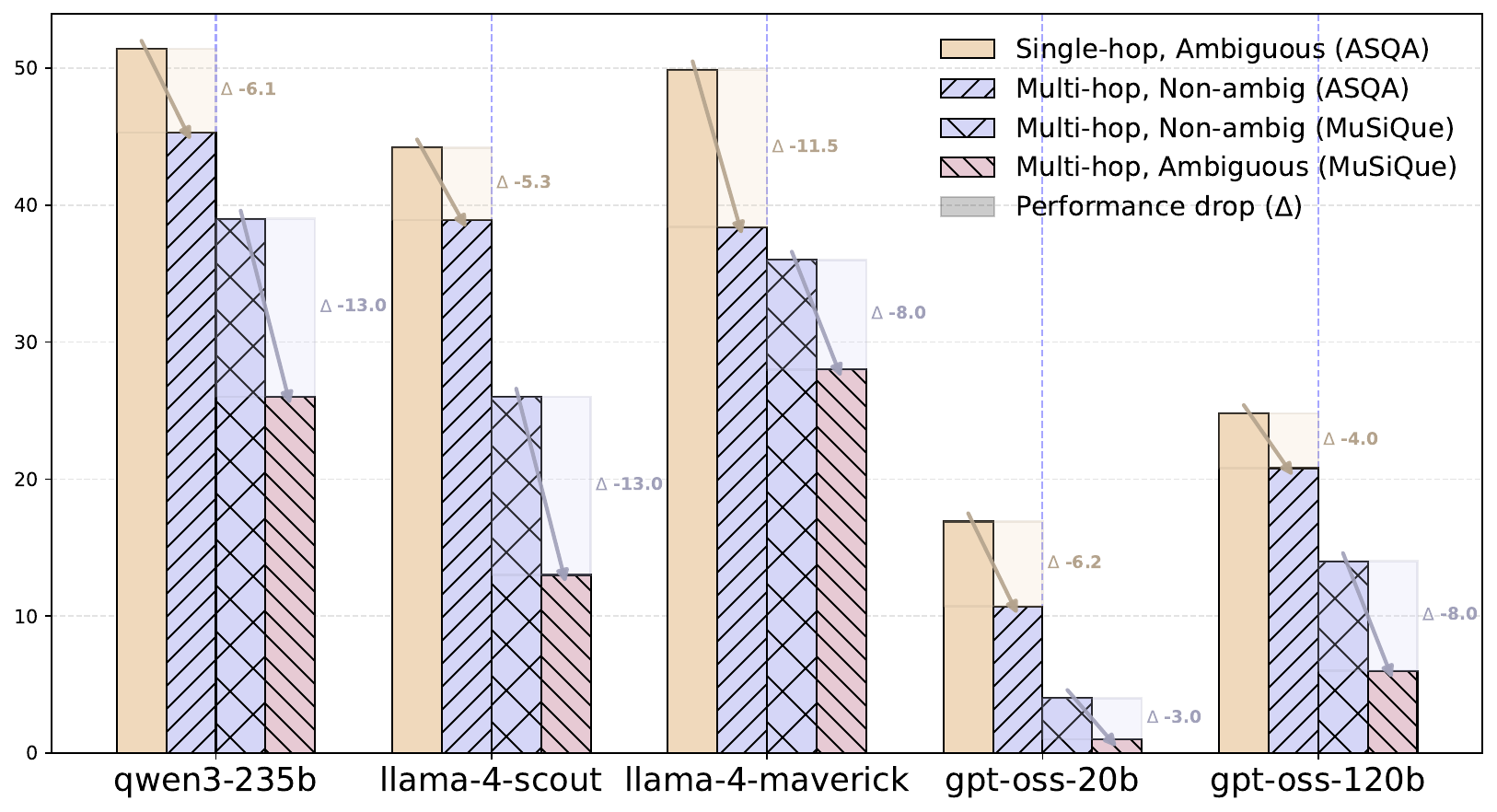}
  \end{subfigure}

  \caption{Multi-hop ambiguity prevalence (top) and performance drops (bottom).}
  \label{fig:challenge_compact}
  \vspace{-6.5mm}
\end{figure}

Figure~\ref{fig:challenge2} illustrates the challenge of multi-hop ambiguous QA. In \textit{``What is the best-selling \textbf{pickup} historically sold by the company that manufactures the `Mustang'?''}, ambiguity arises from the interaction of \textit{Mustang} and \textit{pickup}. Because \textit{pickup} is polysemous (truck vs.\ guitar component), the second-hop ambiguity is \emph{latent} and only surfaces if we keep both \textit{Mustang} interpretations. Current LLMs often commit early to the car reading and prune the valid guitar$\rightarrow$magnetic-pickup branch.
We observe that this layered ambiguity is not a rare edge case. An analysis of real-world user queries from the \textit{lmsys-chat-1m corpus}~\citep{zheng2024lmsyschatm} (Figure~\ref{fig:challenge_compact}, top) reveals that 48.4\% of questions are ambiguous, 17.7\% involve multi-hop reasoning, and 13.3\% overlap. 
Despite this prevalence, empirical results (Figure~\ref{fig:challenge_compact}, bottom) show that datasets like MuSiQue (Multi-hop)~\citep{trivedi2022musique} and ASQA (Ambiguous)~\citep{stelmakh2022asqa} suffer substantial performance drops when these features intersect. This underscores the need for benchmarks that specifically test the ability to hold multiple reasoning paths in superposition.

To address this, we introduce \textbf{M}ulti-hop \textbf{A}mbiguity \textbf{R}easoning \textbf{CH}ain (\textbf{MARCH}), a benchmark designed to evaluate this intersection of ambiguity interpretation and multi-step inference. 
MARCH contains \textbf{2,209} ambiguous multi-hop questions derived from MuSiQue, each paired with type-specific clarified questions (covering multiple interpretations), interpretation-grounded short answers, supporting evidence passages, and a synthesized long answer.
We construct MARCH from MuSiQue through a rigorous pipeline involving multi-LLM verification to ensure quality. To address concerns about LLM-generated data quality, we validate a stratified sample with five human annotators, confirming high long-answer validity (over 90\% integrate all interpretations) and strong inter-annotator agreement (Fleiss' $\kappa$ up to 0.95). 
Our experiments with MARCH reveal that even state-of-the-art models struggle to resolve these layered ambiguities, often producing incomplete or one-sided answers.

To overcome this, we propose \textbf{CLARION} (\textbf{CL}arifying \textbf{A}mbiguity with a \textbf{R}easoning and \textbf{I}nstructi\textbf{ON}), a two-stage agentic framework that decouples \textit{ambiguity planning} from \textit{evidence retrieval}. By explicitly mapping out diverging interpretations via a Planning Agent \textit{before} acting, CLARION prevents the premature pruning of latent branches. Empirical results demonstrate that CLARION significantly outperforms standard baselines, validating the necessity of separating ambiguity resolution from the retrieval loop. 
To isolate the source of difficulty, we also run the baselines on single-hop ambiguity and standard multi-hop datasets, where modern baselines are often reasonably capable in each setting alone. We then see this capability fail on MARCH, where early interpretation choices lock in bridge entities and make ambiguity latent and path-dependent across hops.


\section{Multi-Hop Ambiguous QA}

We define multi-hop ambiguous QA as a task where a single input query triggers multiple valid reasoning chains, requiring the system to resolve uncertainties that determine the trajectory of multi-step inference. In multi-hop ambiguous QA, each valid interpretation dictates a unique sequence of intermediate decomposition steps (e.g., identifying different bridge entities), which in turn necessitates retrieving disjoint sets of evidence documents. Consequently, failing to resolve ambiguity at the initial or intermediate hops leads to a cascading failure, where the reasoning agent pursues an irrelevant trajectory and cannot recover the correct final answer.
To systematically analyze these challenges, we extend the standard ambiguity taxonomy to the multi-hop setting. 
Following prior ambiguity definitions~\citep{tang2025clarifying, tanjim2025disambiguation}, we extend the taxonomy to multi-hop QA and group ambiguous questions into three types---\textit{semantic}, \textit{syntactic}, and \textit{constraint}  (Table~\ref{tab:ambig-taxonomy}).

\begin{figure*}[h] 
    \centering
    \includegraphics[width=0.96\linewidth]{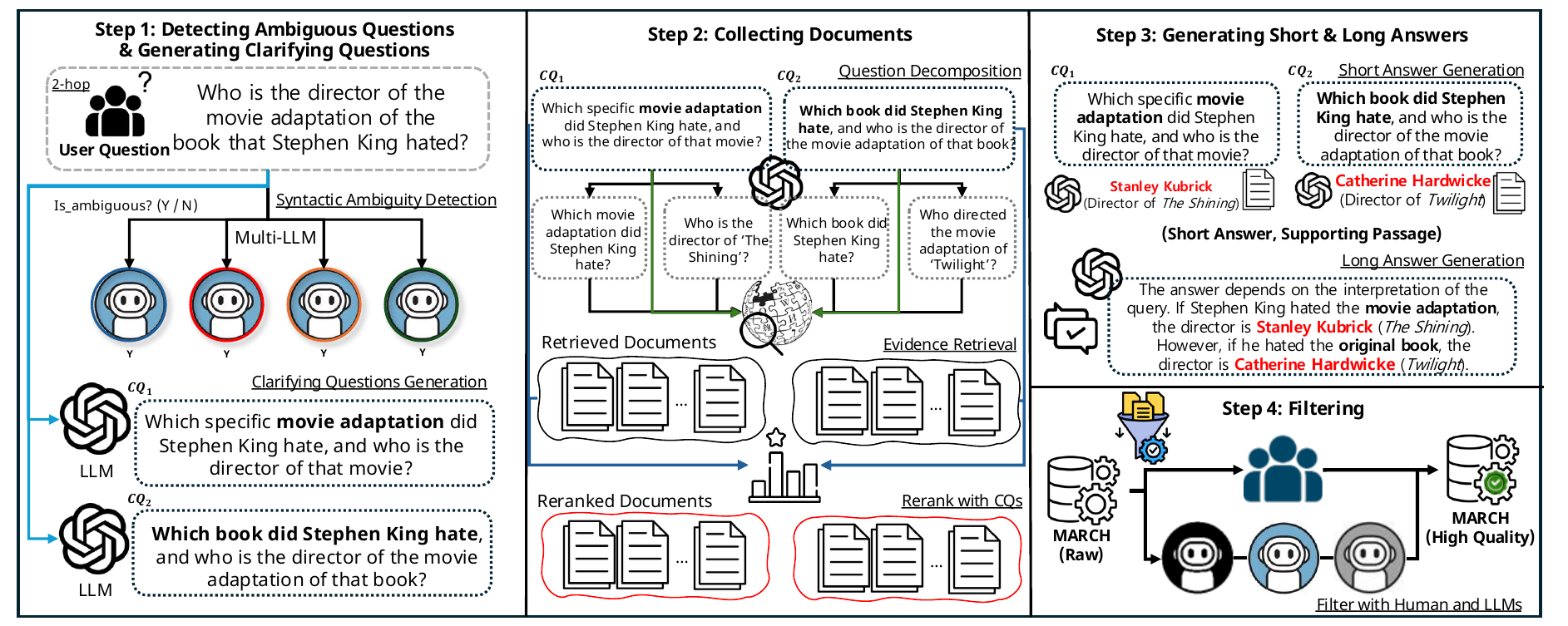}
    \vspace{-2mm}
    \caption{Overview of the four-stage MARCH dataset construction pipeline.}
    \label{fig:dataset_generation}
    \vspace{-5mm}
\end{figure*}

\newcolumntype{L}{>{\RaggedRight\arraybackslash\bfseries}p{55pt}}
\newcolumntype{Y}{>{\RaggedRight\arraybackslash}X}

\begin{table}[t]
\centering
\scriptsize
\setlength{\tabcolsep}{3pt}
\renewcommand{\arraystretch}{1.1}

\begin{tabularx}{\columnwidth}{@{} L Y @{}}
\toprule
\makecell[tl]{\textbf{Type} \\[-0.3ex] \textbf{(LLM Action)}} & \textbf{Definition \& Typical Cues} \\
\midrule
\makecell[tl]{\textbf{Semantic} \\[-0.3ex] (Interpret)} & Homonyms\slash aliases or entity-name collisions (one name, multiple entities); cues: homonyms\slash aliases, acronym collisions, entity-name clashes. \\
\makecell[tl]{\textbf{Syntactic} \\[-0.3ex] (Resolve)} & Multiple valid parses of the same query; cues: pronouns, ellipsis, PP attachment, coordination, quantifier scope. \\
\addlinespace[1ex]
\makecell[tl]{\textbf{Constraint} \\[-0.3ex] (Generalize)} & Over-specific query where a broader related query better matches intent; cues: comparatives\slash superlatives, vague heads, ``overview vs.\ details''. \\
\addlinespace[1ex]

\bottomrule
\end{tabularx}
\vspace{-2mm}
\caption{Taxonomy of multi-hop ambiguity in QA, paired with an LLM action and typical detection cues.}
\vspace{-4.5mm}
\label{tab:ambig-taxonomy}
\end{table}
\vspace{-2mm}
\paragraph{Multi-hop Semantic Ambiguity (Entity-Driven Divergence).}
Semantic ambiguity arises when a mention can map to multiple entities/concepts (e.g., homonymy/entity collision), yielding disjoint evidence trails; choosing the wrong entity invalidates downstream hops. For the example in Figure~\ref{fig:challenge2}: \textit{``What is the best-selling pickup sold by the manufacturer of the `Mustang'?''} branches by the bridge entity:
\textbf{Car:} Mustang$\!\rightarrow$Ford$\!\rightarrow$vehicle sales$\!\rightarrow$F-Series \;\;vs.\;\;
\textbf{Guitar:} Mustang$\!\rightarrow$Fender$\!\rightarrow$pickup types$\!\rightarrow$Single-coil.
Thus, the system must \textsc{Interpret} the mention to select the intended bridge.

\vspace{-2mm}
\paragraph{Multi-hop Syntactic Ambiguity (Structure-Driven Branching).}
Syntactic ambiguity occurs when multiple valid parses induce different inter-hop dependencies, changing which intermediate evidence is needed. For \textit{``What is the model of the telescope the detective saw the suspect with?''},
\textbf{Instrumental:} detective used the telescope$\!\rightarrow$find equipment$\!\rightarrow$model \;\;vs.\;\;
\textbf{Attributive:} suspect had the telescope$\!\rightarrow$find possession$\!\rightarrow$model.
The system must \textsc{Resolve} the parse to construct the correct decomposition plan.

\vspace{-2mm}
\paragraph{Multi-hop Constraint Ambiguity (Scope-Driven Pruning).}
Constraint ambiguity occurs when an over-specific modifier is unnecessary or mismatched with how evidence is written, causing a valid chain to be pruned early. Example:
\textit{``What is the capital of the country where the \underline{highest mountain in Europe} is located?''}
Many sources disagree on whether the highest mountain in Europe is \textit{Mount Elbrus} (Caucasus) or \textit{Mont Blanc} (Alps), and some pages simply say ``Europe's highest mountain'' without committing. If the system enforces the modifier literally, it may retrieve only one interpretation and miss the other. A robust strategy is to \textsc{Generalize} (relax) the constraint (e.g., consider both candidates) and then verify the remaining hop (country $\rightarrow$ capital). For more details for this taxonomy, see Appendix~\ref{appen:constraint}.

\section{MARCH: A Benchmark for Multi-Hop Ambiguous QA}
We introduce MARCH, a benchmark designed to evaluate ambiguity resolution and multi-hop reasoning in question answering jointly.
We process ambiguous questions from MuSiQue through four stages to build MARCH: (1) Ambiguity detection and clarification; (2) Document collection; (3) Generation of short answers for each interpretation and a long answer; and (4) Filtering, as in Figure~\ref{fig:dataset_generation}.

\subsection{Dataset Construction}
\label{sec:construction}
We build MARCH from MuSiQue’s validation set and a subset of its training set. Unlike other multi-hop benchmarks~\citep{zhu2024fanoutqa, he2024mintqa} that are narrow in domain or inflate hops with list-style questions (e.g., top-5), MuSiQue enforces connected, dependency-linked reasoning across diverse domains. 
We first filter out questions from MuSiQue that lack ambiguity and retain only those judged as ambiguous by our multi-stage pipeline.
Let $\mathcal{Q}_{\text{base}}$ be the set of base multi-hop questions. We consider three ambiguity types $\mathcal{T}=\{\text{Semantic}, \text{Syntactic}, \text{Constraint}\}$. We use a set of off-the-shelf LLMs as detectors; for a question $q\in\mathcal{Q}_{\text{base}}$, type $t\in\mathcal{T}$, and detector $m$, let $y_{m,t}(q)\in\{0,1\}$ denote whether $q$ is judged ambiguous of type $t$.

\paragraph{Step 1. Detecting Ambiguous Questions \& Generating Clarified Questions.}  
For each question of MuSiQue, we provide definitions of each ambiguity type and ask multiple LLMs to detect type-wise ambiguity. We employ four detectors: \textit{gpt-4.1}~\citep{achiam2023gpt}, \textit{llama-4-maverick}~\citep{meta2025llama4maverick}, \textit{qwen3-235b-a22b}~\citep{yang2025qwen3}, and \textit{claude-sonnet-4}~\citep{anthropic2025claudesonnet4}. We keep a type label only when the detectors are \emph{fully in agreement}.

Let \(\mathcal{M}\) be the detector set (\(|\mathcal{M}|=4\)). For a question \(q\) and type \(t\in\mathcal{T}\), detector \(m\in\mathcal{M}\) outputs \(y_{m,t}(q)\in\{0,1\}\). We define the full-agreement rule:
\[
\phi_t(q) = \mathbb{I}\!\left(
    \frac{1}{|\mathcal{M}|}\sum_{m\in\mathcal{M}} y_{m,t}(q) = 1
\right),
\]

\[
\mathcal{T}(q) = \{\, t \in \mathcal{T} \mid \phi_t(q) = 1 \,\}.
\]

where \(\mathcal{T}(q)\) denotes the ambiguity types assigned to \(q\). Under this rule, we assign a type \(t\) only if all four detectors judge \(q\) ambiguous of type \(t\), reducing single-model bias and yielding high-quality labels.

For each $(q,t)$ with $t\in T(q)$, we use \textit{gpt-4.1} to generate clarified questions that resolve the type-$t$ ambiguity while preserving the user's information need, denoted as $\mathcal{C}(q,t)=\{c_1,\ldots,c_n\}$ with $n\ge 2$. We utilize these clarified questions as retrieval inputs.

\paragraph{Step 2. Collecting Documents.}  
To generate answers for the questions, we require evidence obtained via retrieval over clarified questions. However, clarified questions can remain multi-hop, and questioning with a specific multi-hop form may narrow the search scope and miss relevant documents. To mitigate this, we use \textit{gpt-4.1} to decompose each clarified question \(c\in\mathcal{C}(q,t)\) into atomic sub-questions \(\mathcal{S}(c)=\{s_1,\ldots,s_k\}\). For every \(s\in\mathcal{S}(c)\), we retrieve up to 10 candidate documents from English Wikipedia\footnote{\url{https://dumps.wikimedia.org/}}. We then pool candidates \(\mathcal{D}(c)=\bigcup_{s\in\mathcal{S}(c)} \mathcal{D}(s)\), and if \(|\mathcal{D}(c)|<10\), we additionally perform retrieval process with the clarified question \(c\) itself to back-fill more evidence. Next, we perform embedding-based re-ranking with \textit{Qwen3-8B-Embedding}~\citep{qwen3embedding} by the similarity between the clarified question and document passages. Finally, we sort \(\mathcal{D}(c)\) by this score to prioritize evidence aligned with the clarified interpretation.

\paragraph{Step 3. Generating Short and Long Answers.}  
Given each clarified question \(c\) and its ranked candidate documents \(\mathcal{D}(c)\), we use \textit{gpt-4.1} to produce a short factual answer only when the retrieved evidence clearly supports it; otherwise, we omit the short answer and drop that clarified item. For retained cases, we also record the passage used for generating the short answer. 
Finally, for the original question $\mathcal{Q}_{\text{base}}$, we utilize \textit{gpt-4.1} to write a single-sentence long answer that connects the two short answers into a coherent statement while incorporating interpretations and citations. 
For the detailed prompts, refer to Appendix~\ref{appen:prompt}.

\paragraph{Step 4. Filtering.}  
Before filtering, we ensure that short answers remain concise. If either clarified question has a short answer longer than 10 tokens, we cut it down to a much shorter form with \textit{gpt-4.1}. After that, we remove cases where the two short answers are identical. (We provide representative examples after applying this collision rule in Appendix~\ref{app:short}.)  
After the filtering, the final MARCH dataset consists of \textbf{2,209} examples.
For the final filtering stage, we exclude \textit{gpt-4.1}—already used in Steps 2 and 3—and instead employ \textit{llama-4-maverick}, \textit{qwen3-235b-a22b}, and \textit{claude-sonnet-4}. Each candidate instance, including the question, clarified questions, ambiguity type, supporting passages, short answers, and long answers, is independently checked for alignment in all fields. We retain only those cases where all three models unanimously judged the instance as fully aligned, using the same criteria as our human evaluation protocol.
The upper side of Table~\ref{tab:merged} shows statistics and reports key characteristics of MARCH. MARCH also retains broad topical coverage; Appendix~\ref{app:domain} reports the domain distribution of MARCH.

\newcolumntype{M}[1]{>{\centering\arraybackslash}m{#1}}

\newcommand{\bfix}[1]{\textbf{\hspace{0.12em}#1}}
\begin{table}[!h]
\centering
\scriptsize
\setlength{\tabcolsep}{2.2pt}
\renewcommand{\arraystretch}{1.08}

\begin{tabular*}{\columnwidth}{@{\extracolsep{\fill}}lrrrr@{}}
\toprule
\textbf{Stage} & \textbf{Sem.} & \textbf{Syn.} & \textbf{Const.} & \textbf{Total} \\
\midrule
MuSiQue (orig.)        & 24{,}834 & 24{,}834 & 24{,}834 & \textemdash \\
After det.\,+clar.     & 9{,}544  & 8{,}642  & 11{,}703 & 29{,}889 \\
After answer gen.      & 7{,}034  & 6{,}675  & 8{,}433  & 22{,}142 \\
Before filtering       & \textbf{1{,}651} & \textbf{1{,}239} & \textbf{1{,}440} & \textbf{4{,}330} \\
After filtering (final)& \textbf{734} & \textbf{739} & \textbf{736} & \textbf{2{,}209} \\
\midrule
Avg.\ hops             & 2.44 & 2.95 & 2.11 & \textemdash \\
Avg.\ Question length  & 14.92 & 18.17 & 16.18 & \textemdash \\
\bottomrule
\end{tabular*}

\vspace{1.0mm}

\begin{tabular*}{\columnwidth}{@{\extracolsep{\fill}}p{0.5\columnwidth}rrrr@{}}
\toprule
\textbf{Evaluation protocol} & \textbf{Sem.} & \textbf{Syn.} & \textbf{Const.} & \shortstack{\textbf{Fleiss'}\textbf{$\kappa$}} \\
\midrule
Question is \textit{type}-ambig.?                 & 94.0\% & 94.0\% & 92.0\% & 0.92 \\
Clarifications resolve ambiguity?                 & 95.0\% & 91.0\% & 96.0\% & 0.89 \\
Long answer matches short answers?                & 97.0\% & 97.0\% & 98.0\% & 0.95 \\
\textbf{All fields valid}                         & \textbf{92.0\%} & \textbf{89.0\%} & \textbf{90.0\%} & \textemdash \\
\bottomrule
\end{tabular*}

\vspace{-1.2mm}
\caption{\textbf{Top:} Statistics of the MARCH dataset filtering pipeline for each ambiguity type. \textbf{Bottom:} Human verification results after filtering, with Fleiss' $\kappa$ indicating high inter-annotator agreement for each ambiguity type.}
\label{tab:merged}
\vspace{-5mm}
\end{table}

\subsection{Dataset Analysis and Validation}
\paragraph{Ambiguity Amplifies Reasoning Depth.} 
We measure the difference in average hop counts between questions assigned ambiguous vs. unambiguous labels via multi-LLM consensus on the MuSiQue training set. After labeling the training questions, we randomly sample 1,000 from each group and compare their average hop counts.
The average hop count for ambiguous questions is \textbf{\textit{2.441}}, while for unambiguous questions it is \textit{2.074}. This result indicates that ambiguous questions generally involve more hops, making them more challenging and underscoring the importance of addressing them effectively. 

\paragraph{Answer Length Statistics.}
Table~\ref{tab:answer_length} reports token-length percentile statistics for both answer types in \textsc{March}.
Each question contains exactly two short answers, one per interpretation, yielding 4,418 short answers in total.
We enforced a strict brevity constraint during the filtering stage (Step~4): if any short answer exceeded 10 tokens, we explicitly prompted the LLM to condense it, and discarded instances where this failed.
As a result, short answers are highly concise (mean 8.27, median 3.00), confirming that they target precise, extractable spans rather than verbose descriptions.
Long answers, which synthesize both interpretations into a coherent statement, are considerably longer (mean 34.86, median 34.00) yet remain focused, with 90\% of instances falling within 50 tokens (P90).
\begin{table}[h]
\centering
\small
\setlength{\tabcolsep}{3pt}
\begin{tabular}{lrrrrrrr}
\toprule
\textbf{Type} & \textbf{$n$} & \textbf{Mean} & \textbf{Med.} & \textbf{P25} & \textbf{P75} & \textbf{P90} & \textbf{P95} \\
\midrule
Short & 4,418 & 8.27 & 3.00 & 2.00 & 15.00 & 22.00 & 27.00 \\
Long  & 2,209 & 34.86 & 34.00 & 26.00 & 42.00 & 50.00 & 55.60 \\
\bottomrule
\end{tabular}
\caption{Token-length statistics for short and long answers in \textsc{March}. Each question has exactly two short answers, one per interpretation.}
\label{tab:answer_length}
\end{table}

\paragraph{Dataset Quality Assessment.}  
To assess dataset quality, we employ five annotators and use a majority-vote scheme.
We first sample 20 instances from the final MARCH dataset per ambiguity type (60 total) and obtain binary (\textsc{Yes}/\textsc{No}) judgments for each item in our protocol, yielding 300 judgments in total (5 annotators $\times$ 60 items). 
As shown in the lower part of Table~\ref{tab:merged}, annotators evaluate whether (i) the question exhibits the specified ambiguity, (ii) the clarified question resolves the original ambiguity, and (iii) the generated long answer contains the corresponding short answers. 
The results demonstrate not only high validity scores (e.g., $>$90\% for all types) but also strong inter-annotator agreement, with Fleiss' Kappa scores of \textbf{0.92} for ambiguity detection, \textbf{0.89} for clarification quality, and \textbf{0.95} for answer consistency. These results confirm the reliability of our automated pipeline in producing high-quality, ambiguous multi-hop questions. 
Please refer to Appendix~\ref{appen:inter} for annotator details, labeling guidelines, and inter-annotator agreement statistics.

\begin{figure*}[!ht] 
    \centering
    \includegraphics[width=0.95\linewidth]{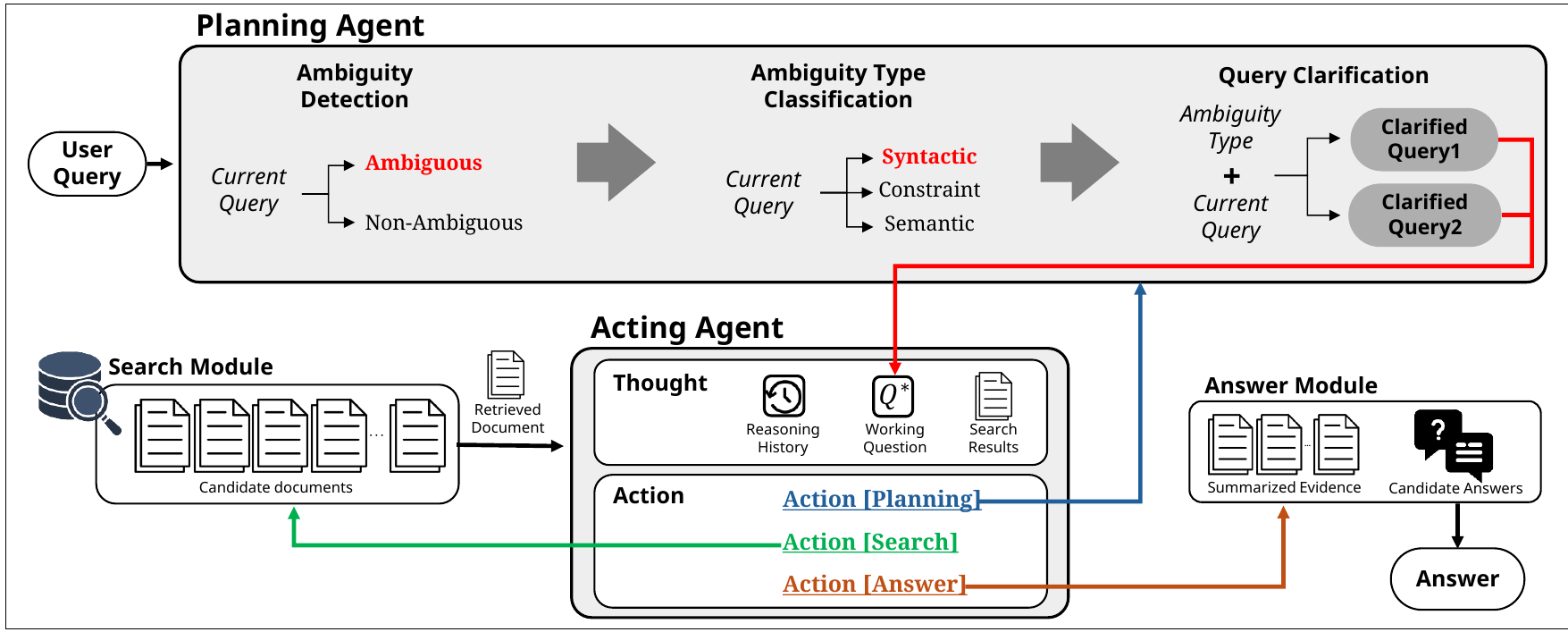}
    \vspace{-1.5mm}
    \caption{Overview of our CLARION framework. A \textit{Planning Agent} resolves ambiguity, and an \textit{Acting Agent} executes a ReAct loop to generate the final answer.}
    \vspace{-5mm}
    \label{fig:method}
\end{figure*}

\vspace{-2mm}
\section{CLARION: An Agentic Framework for Multi-hop Ambiguous QA}
\label{sec:method}
To address multi-hop ambiguous QA, we propose CLarifying Ambiguity with a Reasoning and InstructiON (\textbf{CLARION}), a two-stage agentic framework. As in the example of Figure~\ref{fig:challenge2}, multi-hop ambiguity is often latent; the second hop's ambiguity (``pickup'') only surfaces if the first hop (``Mustang'') is not prematurely resolved. Standard retrieval methods typically fail by committing to a single dominant intent (e.g., Ford cars), thereby pruning alternative paths (e.g., Fender guitars). CLARION overcomes this by explicitly decoupling \textbf{Planning} from \textbf{Acting}, as outlined in Figure~\ref{fig:method}.

\begin{table*}[t]
\centering
\small
\setlength{\tabcolsep}{5pt}
\renewcommand{\arraystretch}{1.1}
\resizebox{0.92\textwidth}{!}{%
\begin{tabular}{llcccc}
\toprule
\textbf{Model} & \textbf{Method} &
\textbf{STR-EM} & \textbf{Disambig-F1)} &
\textbf{Avg} & \textbf{LLM-Judge}\\
\midrule
\multirow{9}{*}{\textit{Qwen3-235b}}
 & No Retrieval                          & 20.98 & 21.19 & 21.09 & \underline{3.083}\\
 & CoT                                   & 21.51 & 22.32 & 21.91 & 2.897\\
 & NaiveRAG                              & 25.10 & \underline{26.20} & 25.65 & 2.752\\
 & CoT w/ RAG                            & 25.63 & 26.61 & 26.12 & 2.947\\
 & DIVA                                  & \underline{28.82} & 22.73 & \underline{25.78} & 3.015\\
 & ReAct                                 & 20.98 & 21.00 & 20.99 & 2.832\\
 & CLARION (Ours)               & \textbf{38.73} & \textbf{28.38} & \textbf{33.56} & \textbf{3.474}\\
 & ~~- w/o clarification                   & 25.10 & 25.56 & 25.33 & 2.922\\
 & ~~- w/o clarification \& detection      & 22.94 & 24.02 & 23.48 & 2.782\\
\midrule
\multirow{9}{*}{\textit{Gemini-2.5}}
 & No Retrieval                          & 15.59 & 20.10 & 17.85 & 2.307\\
 & CoT                                   & 16.32 & 17.52 & 16.92 & 2.258\\
 & NaiveRAG                              & 22.16 & \textbf{28.63} & \underline{25.40} & 2.297\\
 & CoT w/ RAG                            & 23.15 & 27.31 & 25.23 & 2.373\\
 & DIVA                                  & 18.82 & 20.29 & 19.56 & 2.303\\
 & ReAct                                 & 21.32 & 22.37 & 21.84 & 2.428\\
 & CLARION (Ours)               & \textbf{29.12} & \underline{26.30} & \textbf{27.71} & \textbf{2.752}\\
 & ~~- w/o clarification                   & \underline{24.12} & 22.54 & 23.33 & \underline{2.609}\\
 & ~~- w/o clarification \& detection      & 23.82 & 22.04 & 22.93 & 2.573\\
\midrule
\multirow{9}{*}{\textit{DeepSeek-v3.1}}
 & No Retrieval                          & 17.75 & 18.72 & 18.24 & 2.683\\
 & CoT                                   & 19.80 & 22.12 & 20.96 & 2.512\\
 & NaiveRAG                              & 20.20 & \underline{25.03} & 22.62 & 2.084\\
 & CoT w/ RAG                            & 21.33 & 23.18 & 22.25 & 2.632\\
 & DIVA                                  & 18.82 & 20.66 & 19.74 & 2.636\\
 & ReAct                                 & 23.17 & 24.78 & 23.97 & 2.723\\
 & CLARION (Ours)               & \textbf{31.47} & \textbf{27.03} & \textbf{29.25} & \textbf{3.042}\\
 & ~~- w/o clarification                   & 23.63 & 22.99 & 23.31 & \underline{2.927}\\
 & ~~- w/o clarification \& detection      & \underline{24.51} & 24.31 & \underline{24.41} & 2.906\\
\midrule
\textit{Human} &     & 73.00 & 62.00 & 67.50 & {--}\\
\bottomrule
\end{tabular}}
\vspace{-2mm}
\caption{\textbf{MARCH} benchmark results. Metrics are scaled to percentages. \textbf{Bold} = best; \underline{underline} = second best.}
\label{tab:main}
\vspace{-5mm}
\end{table*}

\vspace{-2mm}
\paragraph{Planning Agent.} 
The \textit{Planning Agent} serves as a planning module that analyzes the input question before any retrieval or answering. It performs three sequential operations: (1) \textbf{Ambiguity Detection}: the agent determines whether the question contains ambiguity. If the question is unambiguous, it is immediately passed to the \textit{Acting Agent}. (2) \textbf{Ambiguity Type Classification}: if ambiguity is detected, the question is categorized into one of three predefined types: \textit{Syntactic}, \textit{Constraint}, or \textit{Semantic}. (3) \textbf{Question Clarification}: based on the detected type, the agent rewrites the original question into clarified variants that resolve the ambiguity while preserving the information needed.
These clarified questions constitute the execution plan for downstream reasoning.

\paragraph{Acting Agent.}
The \textit{Acting Agent} executes the reasoning plan through a \textit{ReAct-style prompting}~\citep{yao2023react} scheme, unfolding in a \textbf{Thought $\rightarrow$ Action $\rightarrow$ Observation} loop. 
Unlike naive ReAct, which typically follows a single reasoning trajectory whose early assumptions steer subsequent retrieval, CLARION separates interpretation enumeration from evidence gathering. The Planning Agent produces an initial set of clarified interpretations, and the Acting Agent runs a ReAct loop that retrieves and reasons for each interpretation before synthesizing the final long-form answer. Concretely, the Acting Agent maintains an interpretation set and collects evidence per interpretation, and is instructed to produce a final answer that explicitly covers all interpretations (or as many as possible under the iteration budget).
At each iteration, the agent selects one of three actions: (1) \textbf{Search}: retrieve external documents when additional evidence is needed; (2) \textbf{Planning}: refine or expand the current interpretation set when the current plan is insufficient---for instance, when the agent observes evidence inconsistencies such as an inability to ground an interpretation, a broken hop-to-hop connection, or insufficient branch coverage. This \textit{Planning} action serves as a built-in \textbf{self-correction mechanism}: even if the initial Planning Agent partially misses or misclassifies an ambiguity, the Acting Agent can re-invoke planning to expand or correct the interpretation set and re-initiate retrieval, preventing errors from propagating unchecked through the reasoning chain; (3) \textbf{Answer}: synthesize the final output once enough evidence has been gathered.
To ensure reliable parsing and automated execution, all actions generated by the agent must be in JSON format. Furthermore, to prevent infinite loops and ensure computational tractability, the ReAct prompting is limited to a maximum of five iterations. If this limit is reached without a resolution, the agent is compelled to execute the \textit{Answer} action, formulating the best possible response based on the information gathered thus far. We provide implementation details(models, decoding, retrieval hyperparameters) and the full prompt templates used at each stage in Appendix~\ref{appen:imple} and Appendix~\ref{appen:prompt}, respectively.

\section{Experiments and Results} 

\paragraph{Models.}
We evaluate three state-of-the-art LLMs widely used in real-world scenarios for MARCH: \textit{qwen3-235b-a22b-2507}~\citep{yang2025qwen3}, \textit{gemini-2.5-flash}~\citep{comanici2025gemini}, and \textit{deepseek-chat-v3.1}~\citep{liu2024deepseek}.
To ensure fair comparison, all systems utilize the same retriever (based on \textit{qwen3-embedding-8b}) and identical retrieval hyperparameters.

\vspace{-2mm}
\paragraph{Evaluation metrics.} 
We evaluate performance using three metrics:
(1) \textbf{STR-EM (Strict Exact Match)}: the percentage of gold short answers that appear in the generated long answer after normalization.
(2) \textbf{Disambig-F1}: We use a frozen extractive QA model as an evaluator.
Given the model-generated long answer, we treat it as the context and, for each gold clarified question (one per interpretation), the QA model extracts a short answer span.
We then compute token-level F1 against the corresponding gold short answer and average over all interpretations.
(3) \textbf{LLM-as-a-Judge}: a GPT-4-based judge scores the long answer from 0--5 on \textit{Relevance, Faithfulness, Informativeness,} and \textit{Correctness}; we report the average score (validated in Appendix~\ref{app:val}).
This evaluation framework mirrors the philosophy of structured, sub-factor evaluation. Rather than assessing the long answer holistically, \textbf{STR-EM} and \textbf{Disambig-F1} explicitly decompose evaluation into per-interpretation short answers---one for each valid reading of the ambiguous query. This design achieves the same granular diagnostic insight as a JSON-structured evaluation approach, directly revealing which specific semantic branches a model correctly covers and which it misses, without requiring an additional structured gold format.

\vspace{-2mm}
\paragraph{Baselines.} 
Given that multi-hop ambiguous QA inherently requires external knowledge, we employ search-based baselines for all experiments.
\textbf{(1) No Retrieval}: LLM-only inference without any external context.
\textbf{(2) CoT}~\citep{wei2022chain}: A standard Chain-Of-Thought prompting without retrieval.
\textbf{(3) NaiveRAG}~\citep{lewis2020retrieval}: A standard retrieve-then-read pipeline that retrieves top-$k$ passages using the original question. 
\textbf{(4) CoT with RAG}~\citep{wei2022chain,lewis2020retrieval}: CoT prompting augmented with retrieved documents. 
\textbf{(5) DIVA}~\citep{in-etal-2025-diversify}: 
A \emph{diversify–verify–adapt} framework designed for ambiguous QA. It diversifies the query into multiple interpretations, verifies evidence for each (labeling passages as Useful/Partially Useful/Useless), and adapts its answering strategy based on evidence sufficiency.
\textbf{(6) ReAct}~\citep{yao2023react}: 
An agentic baseline that dynamically interleaves reasoning and retrieval. It generates \textit{Thoughts} to plan, executes retrieval \textit{Actions}, and uses \textit{Observations} to iteratively refine its path.
\vspace{-2mm}
\paragraph{Main Results and Discussion}
As shown in Table~\ref{tab:main}, CLARION consistently outperforms all baselines across all models and metrics. Notably, on \textit{Disambig-F1}—which is most sensitive to the "missing branch" failure mode—CLARION achieves substantial gains. 
This confirms that CLARION uncovers and traverses latent reasoning paths of multi-hop QA that standard methods often prune. 


We attribute this failure to the path-dependent nature of multi-hop ambiguity: committing to an early interpretation conditions downstream decomposition and retrieval, which can prevent alternative later-hop meanings from even surfacing. This error propagation is particularly severe in multi-hop settings because each hop's output serves as the input to the next---a misresolved ambiguity at hop~1 fixes an incorrect bridge entity, which steers hop~2 retrieval toward an irrelevant trajectory, and so on, compounding errors across the entire reasoning chain. 

In particular, the comparison with ReAct underscores the critical role of explicit disambiguation. While ReAct dynamically retrieves information, it often prunes alternative interpretations early by following a single dominant reasoning trace, after which subsequent retrieval further reinforces that commitment. A similar failure arises in standard RAG-based baselines. Retrieving passages with the question over-focuses on the most frequent interpretation and under-covers evidence for other branches, especially when different interpretations induce disjoint bridge entities and documents. In contrast, CLARION's \textit{Planning Agent} forces the exploration of divergent paths \textit{before} retrieval, enabling interpretation-specific evidence collection and preventing premature commitment.
Ablation studies further confirm our design choices. Removing the \emph{Clarification} module causes the largest performance drop, demonstrating that without explicit query rewriting, even agentic systems fail to capture the user's multi-faceted intent.


\paragraph{Human Performance.}
To contextualize the difficulty of \textsc{March}, we evaluate human performance on a sampled subset of 60 questions (20 per ambiguity type). Two graduate-level annotators fluent in English answered each question with unrestricted web search access. The annotators achieved an average STR-EM of 73.0 and Disambig-F1 of 62.0, with strong inter-annotator agreement (Cohen's $\kappa = 0.89$). The substantial gap between human performance and the best-performing system (CLARION: STR-EM 38.73, Disambig-F1 28.38) confirms that \textsc{March} poses a significant and unsolved challenge for current reasoning systems, leaving ample room for future progress.


 \vspace{-2mm}
\newcommand{\cmark}{\textcolor{red}{\ding{51}}}
\newcommand{\xmark}{\ding{55}}

\begin{table*}[t]
\centering
\scriptsize
\setlength{\tabcolsep}{4pt}
\renewcommand{\arraystretch}{1.15}

\begin{tabular*}{\textwidth}{@{\extracolsep{\fill}}
    l r >{\centering\arraybackslash}p{0.38\textwidth} c c c c @{}}
\toprule
\textbf{Benchmark} & \textbf{Scale} & \textbf{Tasks} &
\makecell[c]{\textbf{Multi-}\\\textbf{hop?}} &
\makecell[c]{\textbf{Short}\\\textbf{Ans.?}} &
\makecell[c]{\textbf{Long}\\\textbf{Ans.?}} &
\makecell[c]{\textbf{Ambig.}\\\textbf{Type}\\\textbf{Diversity?}} \\
\midrule
AmbigQA     & 14{,}042 & Ambiguous QA & \xmark & \cmark & \xmark & \xmark \\
CAMBIGNQ    & 5{,}653  & Ambiguity Detection; Clarifying Question Generation
                     & \xmark & \cmark & \xmark & \xmark \\
CondAmbigQA & 200      & Conditional Ambiguous QA
                     & \xmark & \cmark & \xmark & \xmark \\
ASQA        & 5{,}301  & Long-form QA
                     & \xmark & \cmark & \cmark & \xmark \\
AmbigDocs   & 36{,}098 & Ambiguous QA
                     & \cmark & \cmark & \xmark & \xmark \\
DeepAmbigQA & 3{,}600  & Multi-hop QA; Answer Completeness; Name Ambiguity
                     & \cmark & \cmark & \xmark & \xmark \\
\midrule
\textbf{MARCH (Ours)} & \textbf{2{,}209} &
\makecell[c]{\textbf{Multi-hop Ambiguity Detection}\\
            \textbf{Multi-hop Clarifying Question Generation}\\
            \textbf{Multi-hop Long-form QA}}
                     & \cmark & \cmark & \cmark & \cmark \\
\bottomrule
\end{tabular*}
\vspace{-2mm}
\caption{Comparison of ambiguous QA benchmarks and \textbf{MARCH}.}
\label{tab:MARCH-comparison}
\vspace{-4.5mm}
\end{table*}

\paragraph{Performance by Ambiguity Type.}
\label{app:typeperformance}



\begin{table}[!h]
\centering
\small
\setlength{\tabcolsep}{3pt}
\renewcommand{\arraystretch}{1.15}

\begin{tabularx}{\linewidth}{
    >{\raggedright\arraybackslash}p{0.25\linewidth} 
    >{\raggedright\arraybackslash}p{0.20\linewidth} 
    *{3}{>{\centering\arraybackslash}X}
}
\toprule
\textbf{Model} & \textbf{Ambiguity Type} &
\textbf{STR-EM} & \textbf{Disambig-F1} &
\textbf{LLM-as-a-Judge} \\
\midrule
\textit{Qwen3-235b} & Syntactic & 35.00 & 25.81 & 3.370 \\
                    & Constraint   & 46.45 & 33.91 & 3.860 \\
                    & Semantic  & 33.90 & 24.86 & 3.303 \\
\midrule
\textit{Gemini-2.5} & Syntactic & 29.67 & 26.53 & 2.652 \\
                    & Constraint   & 32.79 & 28.71 & 3.059 \\
                    & Semantic  & 24.86 & 23.61 & 2.637 \\
\midrule
\textit{DeepSeek-v3.1} & Syntactic & 29.33 & 27.17 & 2.840 \\
                       & Constraint   & 34.43 & 28.78 & 3.316 \\
                       & Semantic  & 30.23 & 25.11 & 2.940 \\
\bottomrule
\end{tabularx}
\vspace{-1.5mm}
\caption{Performance by ambiguity type on \textbf{MARCH}. Values are percentages except LLM-as-a-Judge.}
\label{tab:ambiguity-typeperformance}
\vspace{-2mm}
\end{table}

Table~\ref{tab:ambiguity-typeperformance} reports performance by ambiguity type across LLMs.
We observe a consistent ordering: \textit{Constraint} $>$ \textit{Syntactic} $\approx$ \textit{Semantic}$.$ This gap mainly reflects whether ambiguity preserves or changes the evidence trail.
For Constraint ambiguity, the competing readings often differ only by an over-specific modifier. Relaxing the constraint typically retains the same bridge entities and yields highly overlapping evidence, allowing models to recover the multi-hop chain even after a suboptimal early choice.
In contrast, Syntactic and Semantic ambiguity more directly steer the hop structure and bridge selection, which can split the reasoning process into branch-specific trajectories. Once a model commits early, retrieval and decomposition become path-dependent, reinforcing that trajectory and causing premature pruning of alternatives.

\vspace{-2mm}
\paragraph{Isolating Multi-hop and Ambiguity effects.}
\begin{table}[!h]
\centering
\scriptsize
\setlength{\tabcolsep}{3.5pt}
\renewcommand{\arraystretch}{1.15}

\resizebox{\columnwidth}{!}{%
\begin{tabular}{lcc|cc}
\toprule
\textbf{Method}
& \multicolumn{2}{c}{\textbf{ASQA}}
& \multicolumn{2}{c}{\textbf{MuSiQue}} \\
\cmidrule(lr){2-3}\cmidrule(lr){4-5}
& \textbf{STR-EM} & \textbf{Disambig-F1}
& \textbf{EM} & \textbf{F1} \\
\midrule
NaiveRAG & 75.95 & 38.50 & 6.20 & 51.87 \\
DIVA     & 71.70 & 32.92 & 11.80 & 53.80 \\
ReAct    & \underline{82.34} & \underline{40.83} & \underline{13.47} & \underline{54.27} \\
CLARION  & \textbf{91.18} & \textbf{48.78} & \textbf{17.07} & \textbf{55.60} \\
\bottomrule
\end{tabular}}

\vspace{-2mm}
\caption{ASQA and MuSiQue results averaged over three LLM backbones.}
\label{tab:asqa_musique_overlap}
\end{table}

To isolate the combined challenge of ambiguity interpretation and multi-hop inference, we evaluate NaiveRAG, DIVA, and ReAct, along with our CLARION, on ASQA, MuSiQue, which respectively probe single-hop ambiguity, standard multi-hop reasoning. From the result of Table~\ref{tab:asqa_musique_overlap}, we find that modern RAG and agentic baselines perform reasonably well on ASQA and MuSiQue, and CLARION is competitive on both, indicating robustness to constraint ambiguity and standard multi-hop reasoning in isolation.
We find that this success does not transfer to MARCH, where ambiguity is latent and path-dependent. Early interpretation choices fix bridge entities, steer downstream retrieval, and prune alternative branches. As a result, baselines often over-focus on a single trajectory and mix cross-branch evidence, causing cascading errors across hops. CLARION targets this failure mode by separating interpretation planning from evidence-driven acting, retrieving per interpretation, and enforcing hop-consistent reasoning.

\section{Related Work}

\paragraph{Ambiguity in QA.} 

Ambiguity in open-domain QA arises from polysemy or insufficient context, permitting multiple reasonable interpretations. Systems must typically clarify the user's intent or provide comprehensive answers covering all possibilities. AmbigQA~\citep{min-etal-2020-ambigqa} formalized this problem with a disambiguation dataset, while ASQA~\citep{stelmakh2022asqa} introduced long-form answers to synthesize multiple interpretations. SituatedQA~\citep{zhang2021situatedqa} and AmbigDocs~\citep{lee2024ambigdocs} expanded this scope by incorporating situational dependencies and conflicting evidence, respectively. To address such ambiguity, many approaches employ question clarification~\citep{min-etal-2020-ambigqa, zhang2023clarify, zhang2024modeling} or retrieval-augmented strategies. Recent RAG-based pipelines~\citep{tanjim2025disambiguation} disambiguate before retrieval, while methods like Tree of Clarifications~\citep{kim-etal-2023-tree} use branching retrieval to explore interpretations. Similarly, DIVA~\citep{in-etal-2025-diversify} adopts a diversify–verify–adapt framework to rewrite queries and synthesize answers from diverse evidence. ReAct~\citep{yao2023react} interleaves reasoning and tool-use in a Thought$\rightarrow$Action$\rightarrow$Observation loop, often following a single trajectory. However, these works primarily target single-hop ambiguity and do not address the compounding uncertainty inherent in multi-hop reasoning chains. 
CLARION instead enumerates interpretations first and retrieves/evaluates evidence per interpretation before synthesis, unlike DIVA which optimizes retrieval quality post-hoc without committing to an explicit interpretation set---leaving it susceptible to premature branch pruning in path-dependent multi-hop settings.
\paragraph{Multi-hop QA.} 
Multi-hop QA requires reasoning across multiple documents~\citep{he2024mintqa, zhu2024fanoutqa, tangmultihop}. HotpotQA~\citep{yang2018hotpotqa} targets the retrieval of articles and sentence-level facts, while 2WikiMultihopQA~\citep{ho-etal-2020-constructing} enhances explainability by providing structured evidence and reasoning paths. 
MuSiQue~\citep{trivedi2022musique} mitigates shallow shortcuts found in prior datasets by enforcing connected reasoning through dependent single-hop questions.
Unlike prior ambiguity benchmarks (e.g., AmbigQA, ASQA, CondAmbigQA~\citep{li2025condambigqa}), which target single-hop questions, MARCH targets the intersection of ambiguity and multi-step inference. As summarized in Table~\ref{tab:MARCH-comparison}, MARCH evaluates the entire pipeline of multi-hop ambiguity detection, clarification, and answer generation.

\section{Conclusion}
We introduce MARCH, a benchmark designed to evaluate ambiguity in multi-hop question answering. MARCH consists of 2,209 carefully annotated questions that include type-specific clarifications, evidence-grounded short and long answers. 
Finally, we propose CLARION, an effective solution for MARCH, and we show that robust LLMs struggle when ambiguity and multi-hop reasoning co-occur.
We find failures are largely path-dependent: early interpretation commitments lock in retrieval and trigger cascading multi-hop errors.

\section*{Limitations}
CLARION demonstrates strong performance but introduces additional system complexity due to its multi-agent structure and planning-acting cycle. Although this complexity enables richer ambiguity resolution, future work could explore lighter-weight or more efficient designs---for instance, a hybrid routing strategy that reserves CLARION for complex, ambiguous queries while forwarding simpler ones to a standard single-step pipeline---without sacrificing effectiveness, making deployment and integration even more practical.

Despite CLARION’s effectiveness, MARCH surfaces the ongoing difficulty of achieving complete and faithful resolution for all multi-hop ambiguous queries. Our results showcase both the progress and the remaining gaps in current methods, providing a solid foundation and clear direction for continued innovation in this important area.

\section*{Ethics Statement}
MARCH benchmark is constructed entirely from publicly available data sources (MuSiQue, Wikipedia), ensuring that no personally identifiable or private information is present. We use a multi-LLM consensus pipeline for ambiguity detection and filtering, reducing the risk of individual model bias or hallucination. Expert contributors, with their consent conduct all human annotation, and no unfair labor practices are involved. While our dataset and evaluation pipeline strive to minimize bias, users should be aware that language models may still inherit subtle biases from the underlying data. We encourage responsible use and further analysis of potential risks when applying MARCH or derived models in real-world settings.


\section*{Acknowledgments}
This work was supported by the Institute of Information \& Communications Technology Planning \& Evaluation (IITP) grant funded by the Korea government (MSIT) [RS-2021-II211341, Artificial Intelligence Graduate School Program (Chung-Ang University)] and the National Research Foundation of Korea(NRF) grant funded by the Korea government(MSIT) (RS-2026-25494299).
This research was supported by the Chung-Ang University Graduate Research Scholarship in 2025.

\bibliography{custom}

@inproceedings{ho-etal-2020-constructing,
    title = "Constructing A Multi-hop {QA} Dataset for Comprehensive Evaluation of Reasoning Steps",
    author = "Ho, Xanh  and
      Duong Nguyen, Anh-Khoa  and
      Sugawara, Saku  and
      Aizawa, Akiko",
    editor = "Scott, Donia  and
      Bel, Nuria  and
      Zong, Chengqing",
    booktitle = "Proceedings of the 28th International Conference on Computational Linguistics",
    month = dec,
    year = "2020",
    address = "Barcelona, Spain (Online)",
    publisher = "International Committee on Computational Linguistics",
    url = "https://aclanthology.org/2020.coling-main.580/",
    doi = "10.18653/v1/2020.coling-main.580",
    pages = "6609--6625"
}

@article{trivedi2022musique,
  title={♫ MuSiQue: Multihop Questions via Single-hop Question Composition},
  author={Trivedi, Harsh and Balasubramanian, Niranjan and Khot, Tushar and Sabharwal, Ashish},
  journal={Transactions of the Association for Computational Linguistics},
  volume={10},
  pages={539--554},
  year={2022},
  publisher={MIT Press One Broadway, 12th Floor, Cambridge, Massachusetts 02142, USA~…}
}

@inproceedings{yang2018hotpotqa,
  title={HotpotQA: A Dataset for Diverse, Explainable Multi-hop Question Answering},
  author={Yang, Zhilin and Qi, Peng and Zhang, Saizheng and Bengio, Yoshua and Cohen, William and Salakhutdinov, Ruslan and Manning, Christopher D},
  booktitle={Proceedings of the 2018 Conference on Empirical Methods in Natural Language Processing},
  pages={2369--2380},
  year={2018}
}

@article{wei2022chain,
  title={Chain-of-thought prompting elicits reasoning in large language models},
  author={Wei, Jason and Wang, Xuezhi and Schuurmans, Dale and Bosma, Maarten and Xia, Fei and Chi, Ed and Le, Quoc V and Zhou, Denny and others},
  journal={Advances in neural information processing systems},
  volume={35},
  pages={24824--24837},
  year={2022}
}

@article{qwen3embedding,
  title={Qwen3 Embedding: Advancing Text Embedding and Reranking Through Foundation Models},
  author={Zhang, Yanzhao and Li, Mingxin and Long, Dingkun and Zhang, Xin and Lin, Huan and Yang, Baosong and Xie, Pengjun and Yang, An and Liu, Dayiheng and Lin, Junyang and Huang, Fei and Zhou, Jingren},
  journal={arXiv preprint arXiv:2506.05176},
  year={2025}
}

@inproceedings{
zheng2024lmsyschatm,
title={{LMSYS}-Chat-1M: A Large-Scale Real-World {LLM} Conversation Dataset},
author={Lianmin Zheng and Wei-Lin Chiang and Ying Sheng and Tianle Li and Siyuan Zhuang and Zhanghao Wu and Yonghao Zhuang and Zhuohan Li and Zi Lin and Eric Xing and Joseph E. Gonzalez and Ion Stoica and Hao Zhang},
booktitle={The Twelfth International Conference on Learning Representations},
year={2024},
url={https://openreview.net/forum?id=BOfDKxfwt0}
}

@inproceedings{lee2024ambigdocs,
  title={AmbigDocs: Reasoning across Documents on Different Entities under the Same Name},
  author={Lee, Yoonsang and Ye, Xi and Choi, Eunsol},
  booktitle={First Conference on Language Modeling},
  year={2024}
}

@inproceedings{zhang2021situatedqa,
  title={SITUATEDQA: Incorporating Extra-Linguistic Contexts into QA},
  author={Zhang, Michael JQ and Choi, Eunsol},
  booktitle={2021 Conference on Empirical Methods in Natural Language Processing, EMNLP 2021},
  pages={7371--7387},
  year={2021},
  organization={Association for Computational Linguistics (ACL)}
}

@inproceedings{stelmakh2022asqa,
  title={ASQA: Factoid Questions Meet Long-Form Answers},
  author={Stelmakh, Ivan and Luan, Yi and Dhingra, Bhuwan and Chang, Ming-Wei},
  booktitle={Proceedings of the 2022 Conference on Empirical Methods in Natural Language Processing},
  pages={8273--8288},
  year={2022}
}

@inproceedings{min-etal-2020-ambigqa,
    title = "{A}mbig{QA}: Answering Ambiguous Open-domain Questions",
    author = "Min, Sewon  and
      Michael, Julian  and
      Hajishirzi, Hannaneh  and
      Zettlemoyer, Luke",
    editor = "Webber, Bonnie  and
      Cohn, Trevor  and
      He, Yulan  and
      Liu, Yang",
    booktitle = "Proceedings of the 2020 Conference on Empirical Methods in Natural Language Processing (EMNLP)",
    month = nov,
    year = "2020",
    address = "Online",
    publisher = "Association for Computational Linguistics",
    url = "https://aclanthology.org/2020.emnlp-main.466/",
    doi = "10.18653/v1/2020.emnlp-main.466",
    pages = "5783--5797"
}

@inproceedings{in-etal-2025-diversify,
    title = "Diversify-verify-adapt: Efficient and Robust Retrieval-Augmented Ambiguous Question Answering",
    author = "In, Yeonjun  and
      Kim, Sungchul  and
      Rossi, Ryan A.  and
      Tanjim, Mehrab  and
      Yu, Tong  and
      Sinha, Ritwik  and
      Park, Chanyoung",
    editor = "Chiruzzo, Luis  and
      Ritter, Alan  and
      Wang, Lu",
    booktitle = "Proceedings of the 2025 Conference of the Nations of the Americas Chapter of the Association for Computational Linguistics: Human Language Technologies (Volume 1: Long Papers)",
    month = apr,
    year = "2025",
    address = "Albuquerque, New Mexico",
    publisher = "Association for Computational Linguistics",
    url = "https://aclanthology.org/2025.naacl-long.56/",
    doi = "10.18653/v1/2025.naacl-long.56",
    pages = "1212--1233",
    ISBN = "979-8-89176-189-6"
}

@inproceedings{kim-etal-2023-tree,
    title = "Tree of Clarifications: Answering Ambiguous Questions with Retrieval-Augmented Large Language Models",
    author = "Kim, Gangwoo  and
      Kim, Sungdong  and
      Jeon, Byeongguk  and
      Park, Joonsuk  and
      Kang, Jaewoo",
    editor = "Bouamor, Houda  and
      Pino, Juan  and
      Bali, Kalika",
    booktitle = "Proceedings of the 2023 Conference on Empirical Methods in Natural Language Processing",
    month = dec,
    year = "2023",
    address = "Singapore",
    publisher = "Association for Computational Linguistics",
    url = "https://aclanthology.org/2023.emnlp-main.63/",
    doi = "10.18653/v1/2023.emnlp-main.63",
    pages = "996--1009"
}

@article{tanjim2025disambiguation,
  title={Disambiguation in Conversational Question Answering in the Era of LLM: A Survey},
  author={Tanjim, Md Mehrab and In, Yeonjun and Chen, Xiang and Bursztyn, Victor S and Rossi, Ryan A and Kim, Sungchul and Ren, Guang-Jie and Muppala, Vaishnavi and Jiang, Shun and Kim, Yongsung and others},
  journal={arXiv preprint arXiv:2505.12543},
  year={2025}
}

@inproceedings{yao2023react,
  title={React: Synergizing reasoning and acting in language models},
  author={Yao, Shunyu and Zhao, Jeffrey and Yu, Dian and Du, Nan and Shafran, Izhak and Narasimhan, Karthik and Cao, Yuan},
  booktitle={International Conference on Learning Representations (ICLR)},
  year={2023}
}

@inproceedings{ho2020constructing,
  title={Constructing A Multi-hop QA Dataset for Comprehensive Evaluation of Reasoning Steps},
  author={Ho, Xanh and Nguyen, Anh-Khoa Duong and Sugawara, Saku and Aizawa, Akiko},
  booktitle={Proceedings of the 28th International Conference on Computational Linguistics},
  pages={6609--6625},
  year={2020}
}

@article{li2025condambigqa,
  title={CondAmbigQA: A benchmark and dataset for conditional ambiguous question answering},
  author={Li, Zongxi and Li, Yang and Xie, Haoran and Qin, S Joe},
  journal={arXiv preprint arXiv:2502.01523},
  year={2025}
}

@article{yang2025qwen3,
  title={Qwen3 technical report},
  author={Yang, An and Li, Anfeng and Yang, Baosong and Zhang, Beichen and Hui, Binyuan and Zheng, Bo and Yu, Bowen and Gao, Chang and Huang, Chengen and Lv, Chenxu and others},
  journal={arXiv preprint arXiv:2505.09388},
  year={2025}
}

@article{comanici2025gemini,
  title={Gemini 2.5: Pushing the frontier with advanced reasoning, multimodality, long context, and next generation agentic capabilities},
  author={Comanici, Gheorghe and Bieber, Eric and Schaekermann, Mike and Pasupat, Ice and Sachdeva, Noveen and Dhillon, Inderjit and Blistein, Marcel and Ram, Ori and Zhang, Dan and Rosen, Evan and others},
  journal={arXiv preprint arXiv:2507.06261},
  year={2025}
}

@article{liu2024deepseek,
  title={Deepseek-v3 technical report},
  author={Liu, Aixin and Feng, Bei and Xue, Bing and Wang, Bingxuan and Wu, Bochao and Lu, Chengda and Zhao, Chenggang and Deng, Chengqi and Zhang, Chenyu and Ruan, Chong and others},
  journal={arXiv preprint arXiv:2412.19437},
  year={2024}
}

@article{achiam2023gpt,
  title={Gpt-4 technical report},
  author={Achiam, Josh and Adler, Steven and Agarwal, Sandhini and Ahmad, Lama and Akkaya, Ilge and Aleman, Florencia Leoni and Almeida, Diogo and Altenschmidt, Janko and Altman, Sam and Anadkat, Shyamal and others},
  journal={arXiv preprint arXiv:2303.08774},
  year={2023}
}

@inproceedings{tang2025clarifying,
  title={Clarifying Ambiguities: on the Role of Ambiguity Types in Prompting Methods for Clarification Generation},
  author={Tang, Anfu and Soulier, Laure and Guigue, Vincent},
  booktitle={Proceedings of the 48th International ACM SIGIR Conference on Research and Development in Information Retrieval},
  pages={20--30},
  year={2025}
}

@misc{meta2025llama4maverick,
  title        = {LLaMA 4 Maverick: A 17B‑128E Multimodal Language Model},
  author       = {Meta AI},
  year         = {2025},
  howpublished = {\url{https://ai.meta.com/blog/llama-4-multimodal-intelligence/}},
  note         = {Available from Meta AI Blog, April 5, 2025}
}

@misc{anthropic2025claudesonnet4,
  title        = {Claude Sonnet 4},
  author       = {Anthropic},
  year         = {2025},
  howpublished = {\url{https://www.anthropic.com/claude/sonnet}},
  note         = {Available on Claude Developer Platform / Anthropic site}
}

@article{he2024mintqa,
  title={MINTQA: A Multi-Hop Question Answering Benchmark for Evaluating LLMs on New and Tail Knowledge},
  author={He, Jie and Hu, Nan and Long, Wanqiu and Chen, Jiaoyan and Pan, Jeff Z},
  journal={arXiv preprint arXiv:2412.17032},
  year={2024}
}

@inproceedings{tangmultihop,
  title={MultiHop-RAG: Benchmarking Retrieval-Augmented Generation for Multi-Hop Queries},
  author={Tang, Yixuan and Yang, Yi},
  booktitle={First Conference on Language Modeling},
  year={2024}
}

@inproceedings{zhu2024fanoutqa,
  title={FanOutQA: A Multi-Hop, Multi-Document Question Answering Benchmark for Large Language Models},
  author={Zhu, Andrew and Hwang, Alyssa and Dugan, Liam and Callison-Burch, Chris},
  booktitle={Proceedings of the 62nd Annual Meeting of the Association for Computational Linguistics (Volume 2: Short Papers)},
  pages={18--37},
  year={2024}
}

@inproceedings{zhang2023clarify,
  title={Clarify When Necessary: Resolving Ambiguity Through Interaction with LMs},
  author={Zhang, Michael JQ and Choi, Eunsol},
  booktitle={Findings of the Association for Computational Linguistics: NAACL 2025},
  pages={5526--5543},
  year={2025}
}

@inproceedings{zhang2024modeling,
  title={Modeling Future Conversation Turns to Teach LLMs to Ask Clarifying Questions},
  author={Zhang, Michael JQ and Knox, W Bradley and Choi, Eunsol},
  booktitle={The Thirteenth International Conference on Learning Representations},
  year={2025}
}

@article{lewis2020retrieval,
  title={Retrieval-augmented generation for knowledge-intensive nlp tasks},
  author={Lewis, Patrick and Perez, Ethan and Piktus, Aleksandra and Petroni, Fabio and Karpukhin, Vladimir and Goyal, Naman and K{\"u}ttler, Heinrich and Lewis, Mike and Yih, Wen-tau and Rockt{\"a}schel, Tim and others},
  journal={Advances in neural information processing systems},
  volume={33},
  pages={9459--9474},
  year={2020}
}

\clearpage
\appendix
\section*{Appendix}

\section{The Use of Large Language Models}
We write the manuscript ourselves, and an LLM (ChatGPT-5.2) is used solely for refinement—style, clarity, and grammar. It is not used for ideation or content generation.




\section{Domain Diversity of MARCH} 
\label{app:domain}
We tag the topic of each question using \textit{gpt-oss-120b} and report the distribution in Table~\ref{tab:MARCH-domain}. As shown in Table~\ref{tab:MARCH-domain}, MARCH covers a broad range of subject areas. The most frequent topics include \textit{“History”}, \textit{“Geography\&Places”}, and \textit{“Politics\&Government”}, indicating diverse coverage beyond any single domain.

\begin{table}[h]
\centering
\footnotesize
\setlength{\tabcolsep}{6pt}
\renewcommand{\arraystretch}{1.15}

\rowcolors{2}{white}{black!2}
\begin{tabular}{l c}
\toprule
\rowcolor{black!5}
\textbf{Domain} & \textbf{Ratio} \\
\midrule
Science \& Technology         & 2.79 \\
Math \& Logic                 & 0.07 \\
History                       & \textbf{30.88} \\
Geography \& Places           & \underline{20.44} \\
Politics \& Government        & 10.16 \\
Business \& Economics         & 2.49 \\
Society \& Culture            & 1.76 \\
Arts \& Literature            & 4.32 \\
Entertainment (Film/TV/Games) & 8.59 \\
Music                         & 7.48 \\
Sports                        & 6.44 \\
Religion \& Philosophy        & 2.31 \\
Medicine \& Health            & 0.18 \\
Nature \& Environment         & 1.57 \\
UNKNOWN                       & 0.51 \\

\midrule
\textbf{Total}                & \textbf{100\%} \\
\bottomrule
\end{tabular}
\caption{MARCH domain coverage.}
\label{tab:MARCH-domain}
\end{table}

\section{Details for Human Annotation}
\label{appen:inter}

\subsection{Details about Annotators}
We employ five graduate-level annotators fluent in English for all labeling tasks in our study. Annotators were compensated at a rate of 10 USD per hour. Each annotator received detailed guidelines and example cases before annotation, and ambiguous cases were discussed through controlled calibration sessions. In total, evaluating the 60 sampled instances for our human assessment required approximately 10 hours of annotation effort. Reporting inter-annotator agreement is crucial for assessing the reliability of human judgments. 
Therefore, we compute Fleiss' $\kappa$, average category-wise agreement ($\bar{P}$), strict agreement (all annotators selecting the same label), and majority agreement (at least three annotators agreeing) for both (1) long-answer judgment dimensions (Relevance, Faithfulness, Informativeness, Correctness) and (2) dataset quality evaluation dimensions (Ambiguity, Clarification, Long Answer).
\begin{table}[t]
\centering
\small
\setlength{\tabcolsep}{2.8pt}      
\renewcommand{\arraystretch}{1.1}

\begin{tabular}{@{} lcccc @{}}
\toprule
\textbf{Item} & \textbf{Fleiss' $\kappa$} & $\bar{P}$ &
\makecell[c]{\textbf{Strict}\\\textbf{Agree}} &
\makecell[c]{\textbf{Maj.}\\\textbf{Agree}} \\
\midrule
Relevance       & 1.000  & 1.000 & 1.000 & 1.000 \\
Faithfulness    & 1.000  & 1.000 & 1.000 & 1.000 \\
Informativeness & 0.907  & 0.978 & 0.967 & 0.989 \\
Correctness     & 1.000  & 1.000 & 1.000 & 1.000 \\
Ambiguity       & 0.589  & 0.978 & 0.967 & 0.989 \\
Clarification   & 0.851  & 0.989 & 0.983 & 0.994 \\
Long Answer     & -0.006 & 0.989 & 0.983 & 0.994 \\
\bottomrule
\end{tabular}

\caption{Inter-annotator agreement across long-answer judgments and
dataset quality evaluation.}
\label{tab:iaa}
\end{table}

\subsection{Human Evaluation Protocols}
\label{appen:protocol}

\begin{table*}[h]
\centering
\footnotesize
\setlength{\tabcolsep}{6pt}
\renewcommand{\arraystretch}{1.2}
\begin{tabular}{l p{0.7\linewidth}}
\toprule
\textbf{Criterion} & \textbf{Description} \\
\midrule
Relevance & Does the long answer fully address both clarified queries and include all relevant short answers, without digression? \\
Faithfulness & Is the answer consistent with the intent and facts in the original query, clarified queries, and short answers? \\
Informativeness & Does the answer provide additional useful background, explanations, or actionable guidance to fulfill the user’s needs? \\
Correctness & Are all facts accurate, with no errors or omissions in the key information? \\
\bottomrule
\end{tabular}
\caption{Human evaluation protocol for long answer quality. Used for correlation analysis in Figure~\ref{fig:corr}.}
\label{tab:human-eval-protocol-appen}
\end{table*}
We develop a dedicated human evaluation protocol to systematically assess long answer quality, as in the correlation analysis presented in Figure~\ref{fig:corr}. Annotators rate each answer using the detailed criteria shown in Table~\ref{tab:human-eval-protocol-appen}, with explicit written instructions for each aspect. This protocol is introduced exclusively for the evaluation setup in Figure~\ref{fig:corr}, ensuring that all human judgments are directly comparable with the figure’s correlation metrics.

\begin{table}[th]
\centering
\setlength{\tabcolsep}{6pt}
\renewcommand{\arraystretch}{1.25}

\resizebox{\columnwidth}{!}{%
    \begin{tabular}{lccc}
    \toprule
    \textbf{Method} & \textbf{Qwen3-235b} & \textbf{Gemini-2.5} & \textbf{DeepSeek-v3.1} \\
    \midrule
    No Retrieval & 0.439 & 0.169 & 0.226 \\
    CoT & 1.295 & 0.476 & 0.862 \\
    NaiveRAG & 0.667 & 0.220 & 0.246 \\
    CoT w/ RAG & 1.042 & 0.440 & 0.778 \\
    DIVA & 1.026 & 0.506 & 0.746 \\
    ReAct & 3.820 & 2.290 & 4.167 \\
    CLARION (Ours) & 8.958 & 2.576 & 5.566 \\
    CLARION w/o clarification & 4.968 & 2.629 & 4.863 \\
    CLARION w/o clarification \& detection & 4.443 & 2.567 & 4.116 \\
    \bottomrule
    \end{tabular}%
}
\caption{Latency (s) comparison across models and methods on the MARCH dataset.}
\label{tab:latency_appendix}
\end{table}

\begin{table*}[t]
\centering
\small
\setlength{\tabcolsep}{6pt}
\renewcommand{\arraystretch}{1.25}

\resizebox{0.8\linewidth}{!}{%
\begin{tabular}{llccc}
\toprule
\textbf{Model} & \textbf{Method} &
\multicolumn{3}{c}{\textbf{Short Answer}} \\
\cmidrule(lr){3-5}
& & \textbf{STR-EM} & \textbf{Disambig-F1} & \textbf{Avg} \\
\midrule
\textit{Qwen3-235b-a22b-250} & No Retrieval                        & 70.98 & 36.56 & 53.77 \\
                              & NaiveRAG                            & 83.14 & 41.34 & 62.24 \\
                              & DIVA                                & 73.53 & 37.54 & 55.54 \\
                              & ReAct & 82.31 & 40.75 & 61.53 \\
                              & CLARION (ours)                      & \textbf{92.94} & \textbf{50.33} & \textbf{71.64} \\
                              & CLARION w/o clarification           & \underline{85.69} & \underline{42.58} & \underline{64.13} \\
                              & CLARION w/o clarification \& detection & 84.31 & 42.30 & 63.30 \\
\midrule
\textit{Gemini-2.5-Flash}     & No Retrieval                        & 66.86 & 34.79 & 50.82 \\
                              & NaiveRAG                            & 76.86 & 39.30 & 58.08 \\
                              & DIVA                                & 66.47 & 35.69 & 51.08 \\
                              & ReAct & 81.13 & 41.72 & 61.42 \\
                              & CLARION (ours)                      & \textbf{89.61} & \textbf{48.02} & \textbf{68.81} \\
                              & CLARION w/o clarification           & \underline{88.43} & \underline{46.09} & 67.26 \\
                              & CLARION w/o clarification \& detection & 87.65 & 46.04 & \underline{66.84} \\
\midrule
\textit{DeepSeek-Chat-v3.1}   & No Retrieval                        & 70.59 & 35.57 & 53.08 \\
                              & NaiveRAG                            & 67.84 & 34.85 & 51.34 \\
                              & DIVA                                & 75.10 & 25.54 & 50.32 \\
                              & ReAct & 83.57 & 40.01 & 61.79 \\
                              & CLARION (ours)                      & \textbf{90.98} & \textbf{47.99} & \textbf{69.48} \\
                              & CLARION w/o clarification           & 87.45 & 46.16 & 66.81 \\
                              & CLARION w/o clarification \& detection & \underline{88.82} & \underline{46.23} & \underline{67.53} \\
\bottomrule
\end{tabular}}
\vspace{-2mm}
\caption{ASQA results across methods and models. We report STR-EM / Disambig-F1 in \%. Best per model in \textbf{bold}, second-best \underline{underlined}.}
\label{tab:asqa}
\end{table*}

\begin{table*}[t]
\centering
\small

\begin{tabular}{lcccccc}
\toprule
\multirow{2}{*}{\textbf{Method}} & \multicolumn{2}{c}{\textbf{Qwen3-235B}} & \multicolumn{2}{c}{\textbf{Gemini-2.5-Flash}} & \multicolumn{2}{c}{\textbf{DeepSeek-Chat-V3.1}} \\
\cmidrule(lr){2-3} \cmidrule(lr){4-5} \cmidrule(lr){6-7}
 & \textbf{EM} & \textbf{F1} & \textbf{EM} & \textbf{F1} & \textbf{EM} & \textbf{F1} \\
\midrule
NaiveRAG & 0.040 & 0.558 & 0.102 & 0.520 & 0.044 & 0.478 \\
DIVA & 0.142 & 0.576 & 0.106 & 0.520 & 0.106 & \textbf{0.518} \\
ReAct Only & 0.140 & 0.568 & 0.132 & 0.551 & 0.132 & 0.509 \\
CLARION (ours) & \textbf{0.206} & \textbf{0.594} & \textbf{0.146} & \textbf{0.560} & \textbf{0.160} & 0.514 \\
\bottomrule
\end{tabular}

\caption{MuSiQue results. Comparison of EM and F1 scores across different models and methods.}
\label{tab:musique}
\end{table*}

Table~\ref{tab:iaa} summarizes the results. Overall, annotators exhibit consistently high agreement across evaluation criteria. The relatively low $\kappa$ value for the \textit{Long Answer} quality dimension stems from a well-known prevalence effect: when nearly all annotators overwhelmingly choose the same label (here, ``Yes''), Fleiss' $\kappa$ is distorted downward due to artificially inflated chance agreement. Importantly, the strict and majority agreement rates for this item remain very high, confirming that annotators were indeed consistent and that the low $\kappa$ does \textit{not} reflect genuine disagreement.

\section{Latency Analysis}
\label{appen:latency}

Table~\ref{tab:latency_appendix} presents the end-to-end latency (in seconds) for each method across three LLM backbones on the MARCH dataset. To ensure a fair comparison, all retrieval-augmented and agentic methods utilize the same retriever and identical retrieval hyperparameters. For agentic approaches, latency is strictly bounded by a fixed interaction budget within the acting loop (maximum of five ReAct-style iterations), preventing unbounded tool calls and ensuring predictable runtime.

Overall, methods without retrieval exhibit the lowest latency, whereas iterative agentic methods are the most computationally intensive. Incorporating Chain-of-Thought (CoT) increases latency compared to direct answering due to the generation of intermediate reasoning steps. Standard retrieve-then-read pipelines (NaiveRAG, CoT w/ RAG, DIVA) incur only modest overhead from retrieval, remaining substantially faster than multi-step tool-use frameworks. In contrast, ReAct shows a marked increase in latency as it interleaves reasoning with multiple sequential search steps. CLARION yields the highest latency in most settings, as it (i) executes a dedicated planning stage for ambiguity detection and clarification, and (ii) performs retrieval and reasoning separately for each clarified interpretation, effectively conducting multi-branch evidence gathering prior to synthesis.

The ablation studies highlight the inherent cost–quality trade-off of explicit clarification. Removing the clarification stage reduces latency by approximately half for Qwen3-235b and DeepSeek-v3.1, though the impact is less pronounced for Gemini-2.5. Ultimately, the latency results align with our design philosophy: CLARION deliberately allocates additional computational resources to preserve and explore multiple interpretations—rather than prematurely committing to a single dominant branch—which is indispensable for resolving the latent, path-dependent ambiguity characteristic of multi-hop QA.

\section{LLM-as-a-Judge}
\subsection{Validating the LLM Judge.}
\label{app:val}
\begin{figure}[t]
\centering

\includegraphics[width=\linewidth]{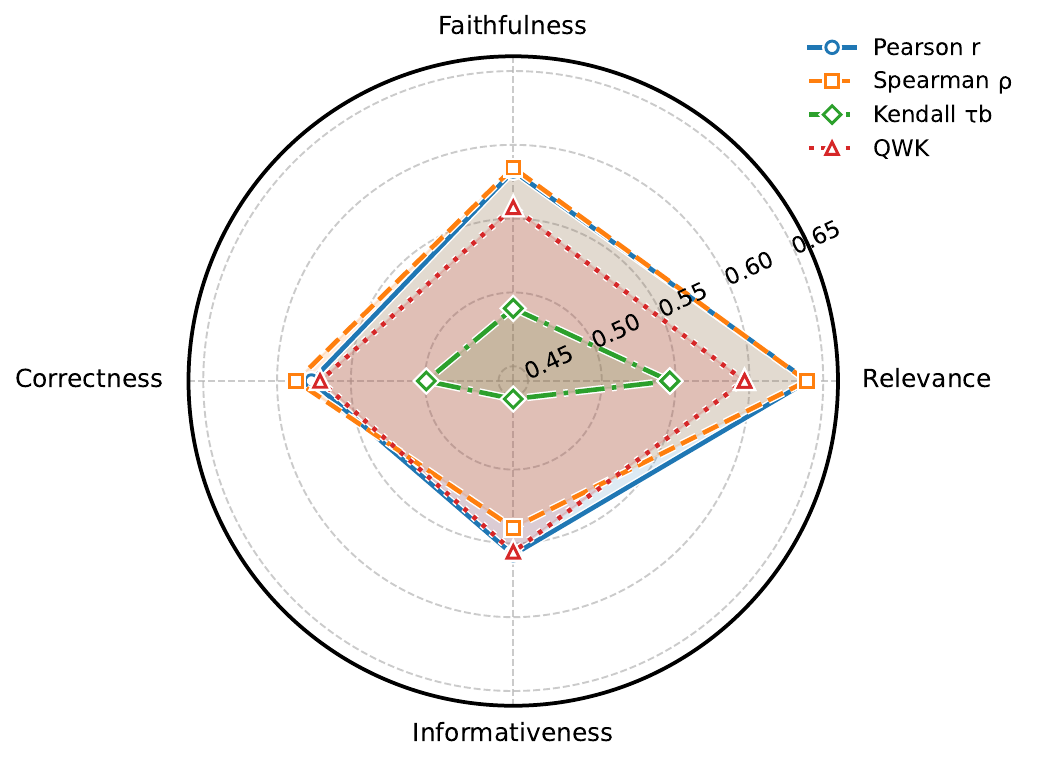}

\caption{Correlation between LLM and human judgments.}
\label{fig:corr}
\end{figure}

We validate the use of the \emph{LLM judge} for our main experiments by comparing its $0$–$5$ scores with human ratings on 300 items across four criteria. We measure (i) linear association with Pearson $r$, (ii) rank consistency with Spearman $\rho$ and Kendall $\tau_b$, and (iii) grade-level agreement on the $0$–$5$ scale via Quadratic Weighted Kappa (QWK). As shown in Figure~\ref{fig:corr}, we observe consistently strong correlations across all families; QWK further indicates grade-aligned agreement, supporting the LLM judge as a valid proxy for our main experiments. For full details on the human evaluation protocol, see Appendix~\ref{appen:protocol}.

\section{Results on Other Related Benchmarks}

\subsection{Results for ASQA Benchmark}
\label{app:asqa}
As shown in Table~\ref{tab:asqa}, in ASQA, our agentic approach consistently delivers the strongest short-answer performance across models. With detection + clarification enabled, CLARION achieves the best per-model averages—e.g., Qwen3-235b-a22b-250: 71.64 vs. 62.24 (NaiveRAG) and 55.54 (DIVA); Gemini-2.5-Flash: 68.81 vs. 58.08 and 51.08; DeepSeek-Chat-v3.1: 69.48 vs. 51.34 and 50.32. Improvements appear in both STR-EM (coverage of gold short answers) and Disambig-F1 (extractability for clarified questions), indicating that explicitly detecting ambiguity and rewriting the query steers retrieval to interpretation-aligned evidence rather than memorized or mixed contexts. Ablations verify the contribution of each component, with the largest drop when clarification is removed—highlighting that planning for ambiguity before acting is crucial even on single-hop-oriented datasets like ASQA. Together, these trends support our claim that agentic planning and acting modules are broadly beneficial beyond MARCH and strengthen answer completeness and precision in ambiguous QA settings.

\subsection{Results for MuSiQue Benchmark}
As shown in Table~\ref{tab:musique}, CLARION consistently outperforms baseline methods on the MuSiQue dataset, which evaluates standard multi-hop reasoning without the specific focus on ambiguity found in MARCH. On average across the three LLM backbones, CLARION achieves an Exact Match (EM) score of 20.6 and an F1 score of 59.4. This represents a clear improvement over the strongest agentic baseline, ReAct (EM: 14.0, F1: 56.8), as well as DIVA (EM: 14.2, F1: 57.6). The results indicate that the benefits of CLARION’s architecture extend beyond ambiguity resolution to general multi-step inference tasks. Even in non-ambiguous multi-hop scenarios, CLARION's ability to decompose complex queries and verify evidence at each hop ensures that the model maintains a coherent trajectory across connected documents.


\subsection{Detailed Score for LLM-as-a-Judge}
\label{app:llmjudge}
As shown in Table~\ref{tab:lj_criteria}, we report the scores of each sub-criterion under the LLM-as-a-Judge evaluation for the baselines and for CLARION. Consistent with our main experiments, CLARION generally achieves higher judge scores across most criteria.

\renewcommand{\dplus}[1]{\,{\scriptsize\textcolor{red}{(+#1)}}}
\renewcommand{\dminus}[1]{\,{\scriptsize\textcolor{blue}{(-#1)}}}
\renewcommand{\GroupHeader}[1]{%
  \rowcolor{black!5}\multicolumn{6}{l}{\textbf{\textit{#1}}}\\[-2pt]
  \cmidrule{1-6}
}

\begin{table*}[t]
\centering
\small
\setlength{\tabcolsep}{6pt}
\renewcommand{\arraystretch}{1.12}
\resizebox{\linewidth}{!}{%
\begin{tabular}{llcccc}
\toprule
\textbf{Method} & \textbf{Model} &
\multicolumn{4}{c}{\textbf{LLM-as-a-Judge (0--5)}} \\
\cmidrule(lr){3-6}
 &  & \textbf{Relevance} & \textbf{Faithfulness} &
      \textbf{Informativeness} & \textbf{Correctness} \\
\midrule

\GroupHeader{LLM-only}
No Retrieval & \textit{Qwen-3-235b}   & 3.324 & 3.147 & 2.931 & 2.931 \\
             & \textit{Gemini-2.5}    & 2.643 & 2.439 & 1.912 & 2.233 \\
             & \textit{DeepSeek-v3.1} & 3.067 & 2.706 & 2.480 & 2.480 \\
\midrule

\GroupHeader{RAG-based baselines}
Naive RAG    & \textit{Qwen-3-235b}   & 2.961 & 2.988 & 2.357 & 2.829 \\
             & \textit{Gemini-2.5}    & 2.525 & 2.620 & 1.643 & 2.543 \\
             & \textit{DeepSeek-v3.1} & 2.325 & 2.408 & 1.516 & 2.259 \\
\cmidrule{1-6}
Diva         & \textit{Qwen-3-235b}   & 3.073 & 3.465 & 2.565 & 3.084 \\
             & \textit{Gemini-2.5}    & 2.531 & 2.694 & 1.727 & 2.382 \\
             & \textit{DeepSeek-v3.1} & 2.918 & 2.851 & 2.292 & 2.596 \\
\midrule

\GroupHeader{CLARION (ours)}
CLARION{w/o clarification \& detection}
             & \textit{Qwen-3-235b}   & 3.057 & 2.957 & 2.496 & 2.696 \\
             & \textit{Gemini-2.5}    & 2.673 & 2.963 & 2.039 & 2.702 \\
             & \textit{DeepSeek-v3.1} & 3.075 & 3.159 & 2.635 & 2.839 \\
\cmidrule{1-6}
CLARION{w/o clarification}
             & \textit{Qwen-3-235b}   & 3.184 & 3.084 & 2.608 & 2.890 \\
             & \textit{Gemini-2.5}    & 2.712 & 2.963 & 2.133 & 2.700 \\
             & \textit{DeepSeek-v3.1} & 3.089 & 3.136 & 2.699 & 2.843 \\
\cmidrule{1-6}
CLARION      & \textit{Qwen-3-235b}   & 3.600 & 3.551 & 3.502 & 3.271 \\
             & \textit{Gemini-2.5}    & 2.843 & 3.106 & 2.302 & 2.824 \\
             & \textit{DeepSeek-v3.1} & 3.228 & 3.177 & 2.943 & 2.882 \\
\bottomrule
\end{tabular}}
\caption{LLM-as-a-Judge sub-criteria. All scores are on a 0--5 scale.}
\label{tab:lj_criteria}
\end{table*}

\section{ReAct Iteration Limit Ablation}
\label{appendix:iteration}

We set the maximum number of ReAct iterations to five based on an empirical ablation varying the limit across $\{1, 3, 5, 7\}$ on Qwen3-235b.
As shown in Table~\ref{tab:iteration_ablation}, performance improves steadily as the iteration limit increases from 1 to 5.
Beyond 5 iterations, however, gains saturate: increasing the limit to 7 yields only negligible improvement ($+$0.10 Avg) while incurring unnecessary computational overhead and latency.
We therefore adopt 5 as the optimal trade-off between performance and efficiency.

\begin{table}[h]
\centering
\small
\setlength{\tabcolsep}{4pt}
\begin{tabular}{ccccc}
\toprule
\textbf{Iter.} & \textbf{STR-EM} & \textbf{Disambig-F1} & \textbf{Avg} & \textbf{LLM-Judge} \\
\midrule
1 & 28.86 & 22.45 & 25.65 & 2.850 \\
3 & 36.82 & 26.91 & 31.86 & 3.315 \\
5 & \underline{38.73} & \underline{28.38} & \underline{33.56} & \textbf{3.474} \\
7 & \textbf{38.81} & \textbf{28.52} & \textbf{33.66} & \underline{3.461} \\
\bottomrule
\end{tabular}
\caption{Impact of ReAct iteration limit on CLARION (Qwen3-235b). Performance saturates beyond 5 iterations with negligible gains. \textbf{Bold} = best; \underline{underline} = second best.}
\label{tab:iteration_ablation}
\end{table}

\section{Case Study}

\subsection{Failure Cases of CLARION}
Despite its strong performance, CLARION still fails on certain ambiguous multi-hop queries. Table~\ref{tab:clarion-failures} shows representative failure cases across all three ambiguity types. In each case, CLARION’s prediction collapses to a single interpretation without resolving ambiguity, so the system generates only one short answer and misses the gold interpretations. This results in complete mismatches (0 on STR-EM and Disambig-F1). These errors illustrate how mis-specified sub-questions or over-broad interpretations derail reasoning and retrieval, leading to a complete mismatch against gold answers.

\subsection{Case Study: A 3-hop Query with Three Interpretations}
Table~\ref{tab:case-3hop-3interp} illustrates how \textbf{CLARION} handles a nominal 3-hop query whose hop-1 is semantically ambiguous. Because \emph{The Birds and the Bees} may refer to different performers/versions, the \textbf{Planning Agent} first detects ambiguity and enumerates three clarified interpretations (I$_1$--I$_3$). The \textbf{Acting Agent} then executes the same 3-hop schema (performer/version $\rightarrow$ birthplace $\rightarrow$ largest annual event in the birthplace) for each interpretation in a \textbf{ReAct-style} loop, carrying entities forward across hops while retrieving evidence step-by-step. Crucially, CLARION enforces \textbf{hop-consistency verification}: the performer/version at H1 must be supported by retrieved evidence, and the event evidence at H3 must be explicitly grounded in the location resolved at H2 before downstream propagation. In this example, I$_1$ passes verification and yields a grounded answer (\emph{Houston Livestock Show and Rodeo}). In contrast, I$_2$ retrieves a plausible event candidate but fails verification due to missing/weak grounding across hops (e.g., unsupported H1 and/or an H2--H3 mismatch), so the Acting Agent triggers \textbf{recovery} via targeted \textbf{re-retrieval} or user confirmation. Finally, I$_3$ correctly blocks execution and requests the missing artist/version, preventing ungrounded multi-hop synthesis.

\newcolumntype{Y}{>{\RaggedRight\arraybackslash}X}  
\newcolumntype{P}[1]{>{\RaggedRight\arraybackslash}p{#1}} 

\begin{table*}[h]
\centering
\small
\renewcommand{\arraystretch}{1.15}
\setlength{\tabcolsep}{5pt}
\begin{tabularx}{\linewidth}{P{0.11\linewidth} P{0.18\linewidth} Y Y P{0.10\linewidth} P{0.10\linewidth}}
\toprule
\textbf{Type} & \textbf{QID} & \textbf{Clarified Query 1} & \textbf{Clarified Query 2} & $\mathrm{SA}_1$ & $\mathrm{SA}_2$ \\
\midrule
Syntactic &
\makecell[tl]{4hop1\_\_8294\\15324\_26424\\\_581618} &
Who founded the chain of music-themed restaurants whose first establishment was located in the birthplace of the person who ejected the Benedictines in 1559? &
Who founded the chain of music-themed restaurants with its first establishment in the place where the person who ejected the Benedictines in 1559 was born? &
Isaac Tigrett & Isaac Tigrett \\
\addlinespace
Constraint &
\makecell[tl]{2hop\_\_100274\\\_14948} &
What latitude marks the northern border of Antarctica? &
At what latitude is the continental boundary of Antarctica defined? &
60° S & 60° S \\
\addlinespace
Semantic &
\makecell[tl]{2hop\_\_725611\\\_52870} &
Which actor from \emph{Michael Collins} appears in \emph{The Phantom Menace}, and which character do they portray? &
In \emph{The Phantom Menace}, which character is played by an actor who was also in \emph{Michael Collins}? &
Liam Neeson & Liam Neeson \\
\bottomrule
\end{tabularx}
\caption{\textbf{Failure cases during MARCH construction due to short-answer shortening collisions.}
After shortening, both short answers collapsed to identical strings, causing removal even though the clarified queries represent distinct interpretations.}
\label{tab:removed-cases}
\end{table*}

\begin{table*}[h]
\centering
\small
\renewcommand{\arraystretch}{1.15}
\setlength{\tabcolsep}{5pt}
\begin{tabularx}{\linewidth}{P{0.12\linewidth} Y Y Y P{0.18\linewidth}}
\toprule
\textbf{Type} & \textbf{Original Query} & \textbf{Predicted Long Answer} & \textbf{Gold Long Answer} & \textbf{Fail Reason} \\
\midrule
Semantic &
What city shares a border with the place where the person who went to the state known for its Mediterranean climate during the gold rush worked? &
Stockton &
Brooklyn and Traverse City share borders with the relevant places. &
Collapsed to a single interpretation; failed to clarify multiple possible places. \\
\addlinespace
Syntactic &
Who won the Indy Car Race in the largest populated city of the state where Yuma's Library District is located? &
Mario Andretti (Phoenix, 1993) &
Álex Palou and Hélio Castroneves &
Relied on historical fact lookup; failed to disambiguate event scope and multiple winners. \\
\addlinespace
Constraint &
Who brought the language Hokkien to the country on the natural boundary between the country that hosted the tournament and the country where A Don is from? &
The Hoklo (Hokkien) people &
Dutch colonial administration and Hokkien-speaking immigrants during Spanish colonization &
Did not resolve broad/general query; simplified to one actor instead of multiple sources. \\
\bottomrule
\end{tabularx}
\caption{\textbf{Representative CLARION failure cases.}
CLARION often fails to clarify ambiguous sub-questions and collapses to a single short answer,
leading to zero scores on both STR-EM and Disambig-F1.}
\label{tab:clarion-failures}
\end{table*}


\begin{table*}[t]
\centering
\scriptsize
\setlength{\tabcolsep}{4.0pt}
\renewcommand{\arraystretch}{1.25}
\setlength{\extrarowheight}{1.2pt}

\begin{tabularx}{\textwidth}{@{}>{\raggedright\arraybackslash}p{0.11\textwidth}YYY@{}}
\toprule
\textbf{Step} &
\textbf{I$_1$: Jewel Akens (1964)} &
\textbf{I$_2$: Dean Martin (claimed)} &
\textbf{I$_3$: Other/unknown artist} \\
\midrule

\multicolumn{4}{@{}>{\raggedright\arraybackslash}p{\textwidth}@{}}{%
\textbf{User query.} \emph{What is the largest annual event held in the birthplace of the performer of
\emph{The Birds and the Bees}?}%
}
\\
\addlinespace[1.1ex]

\multicolumn{4}{@{}>{\raggedright\arraybackslash}p{\textwidth}@{}}{%
\textbf{Detect.} Ambiguous (all detectors: \texttt{Y}): \emph{The Birds and the Bees} can refer to different
performers/versions, so hop-1 branches the entire 3-hop chain.%
}
\\
\addlinespace[0.8ex]
\midrule

\textbf{Clarify} &
Performer/version: Jewel Akens (1964)\newline
Birthplace: Houston, TX
&
Performer/version: Dean Martin (version)\newline
Birthplace: \emph{to verify}
&
Performer/version: \emph{unspecified}\newline
Ask user to specify
\\
\addlinespace[1.2ex]

\textbf{Plan (3-hop)} &
\textbf{H1} identify performer/version\newline
\textbf{H2} retrieve birthplace\newline
\textbf{H3} largest annual event in birthplace
&
\textbf{H1} identify performer/version\newline
\textbf{H2} retrieve birthplace\newline
\textbf{H3} largest annual event in birthplace
&
\textbf{H1} missing $\rightarrow$ cannot execute\newline
\textbf{H2--H3} blocked
\\
\addlinespace[1.2ex]

\textbf{Retrieve} &
\textbf{H1} evidence: \emph{The Birds and the Bees} (1964) performed by Jewel Akens
(doc: \texttt{12676138}).\newline
\textbf{H3} evidence: \emph{Houston Livestock Show and Rodeo} held in Houston
(doc: \texttt{318775}).
&
Candidate \textbf{H3} event: \emph{Pennsic War} (doc: \texttt{535960}).\newline
\textbf{Issue:} H1 (performer/version) is unsupported and H2 (birthplace) is ungrounded,
so the event is not tied to the H2 location (\textbf{H2--H3 mismatch}).
&
No retrieval: artist/version for H1 is missing, so H2--H3 cannot be queried.
\\
\addlinespace[1.2ex]

\textbf{Verify} &
\textbf{Pass:} H1 supported; H2 (Houston) grounds H3 evidence.
&
\textbf{Fail:} hop-consistency break (H1 unsupported and/or H2--H3 mismatch).
&
\textbf{Blocked:} missing H1 prevents verification and downstream hops.
\\
\addlinespace[1.2ex]

\textbf{Output} &
\textsc{Answer:} Houston Livestock Show and Rodeo (with citation).
&
\textsc{Recover:} re-retrieve for H1/H2 or ask user to confirm intended performer.
&
\textsc{Clarify:} ask for the artist/version.
\\
\addlinespace[0.8ex]
\midrule

\multicolumn{4}{@{}>{\raggedright\arraybackslash}p{\textwidth}@{}}{%
\textbf{Final (answer-focused).} Verified answer is \textbf{I$_1$: Houston Livestock Show and Rodeo}.
I$_2$ is not safely answerable without additional grounding; I$_3$ requires user specification.%
}
\\
\bottomrule
\end{tabularx}

\vspace{-1.0mm}
\caption{Case study (qid: \texttt{3hop1\_\_337919\_841757\_11974}): hop-1 semantic ambiguity yields three
interpretations. Each branch follows the same 3-hop schema and is checked by hop-consistency verification
before producing outputs.}
\label{tab:case-3hop-3interp}
\vspace{-2.0mm}
\end{table*}

\section{Implementation Details}
\label{appen:imple}
\begin{table*}[h]
\centering
\footnotesize
\setlength{\tabcolsep}{6pt}
\renewcommand{\arraystretch}{1.15}
\rowcolors{2}{white}{black!2}
\begin{tabular}{l l}
\toprule
\rowcolor{black!5}
\textbf{Item} & \textbf{Value / Setting} \\
\midrule
API & OpenRouter API \\
Detection Model & GPT-4.1, Llama-4-Maverick, Qwen3-235b, Claude-Sonnet-4 \\
Generator Model & GPT-4.1 \\
Filtering Model & Llama-4-Maverick, Qwen3-235b, Claude-Sonnet-4 \\
LLM-as-a-Judge Model & GPT-4.1 \\
Temperature & 0.0 (detection), 0.0 (generation) \\
Max Tokens & 512 (detection), 512 (generation) \\
Evaluation Protocol & See Appendix~\ref{appen:protocol} and Table~\ref{tab:merged} \\
Embedding Model & Qwen3-8B-Embedding \\
Retriever & FAISS \\
Top-k & 10 \\
Agent Max Search Iteration & 5 \\
GPU & RTX 6000 Ada \\
\bottomrule
\end{tabular}
\caption{Implementation details.}
\label{tab:details}
\end{table*} 
As shown in Table~\ref{tab:details}, we report our implementation details for our MARCH construction pipeline and running CLARION.

\paragraph{API Access and Infrastructure.}
All LLM calls were made using the OpenRouter API. Experiments were run on a workstation equipped with an RTX 6000 Ada GPU.

\paragraph{Models.} We use four off-the-shelf LLMs as ambiguity detectors: GPT-4.1 (OpenAI), Llama-4-Maverick (Meta), Qwen3-235b (Alibaba), and Claude-Sonnet-4 (Anthropic). For generating clarified questions and answers, we exclusively use GPT-4.1. We use gpt-4.1 (snapshot: gpt-4.1-2025-04-14) in all experiments to ensure reproducibility. For LLM-as-a-judge filtering, we employ Llama-4-Maverick, Qwen3-235b, and Claude-Sonnet-4. 

\paragraph{Decoding and Prompting.}
All LLM calls (for both detection and generation) were run with temperature set to 0.0 and a maximum token limit of 512. All prompts and task templates are described in detail in Section~\ref{appen:prompt}.

\paragraph{Retrieval Pipeline.}
For evidence retrieval, we use FAISS for fast vector search over Wikipedia passages. Query and document embeddings are computed with the Qwen3-8B-Embedding model. Retrieval is performed with a fixed top-$k$ of 10 per query.

\paragraph{Agentic Reasoning.}
For CLARION, the agent’s maximum search iteration is set to 5. The planning agent performs ambiguity detection, type classification, and clarification as described in Section~\ref{sec:method}; the acting agent executes search and answer steps up to the iteration limit.

\paragraph{Filtering and Evaluation Protocol.}
After answer generation, candidate instances are filtered using three LLMs (excluding GPT-4.1 to prevent overfitting). An instance is retained only if all models unanimously judged every field (question, clarifications, type, evidence, answers) as fully aligned, following the same protocol as human evaluation (see Appendix~\ref{appen:protocol} and Table~\ref{tab:merged} for details).

\paragraph{Ambiguity Taxonomy for Constructing Data}
\label{app:short}

\noindent\textbf{Syntactic:} Two clarified questions differ in grammatical structure (``in the birthplace of'' vs.\ ``in the place where ... was born''), but both shorten to the same short answer.

\noindent\textbf{Constraint:} Two clarified questions ask about different ways of defining Antarctica’s border, yet both reduce to the same numeric answer.

\noindent\textbf{Semantic:} Two clarified questions focus on different semantic aspects (actor identity vs.\ character role), but shortening collapsed both to the same name.

\section{Detection cue for Constraint ambiguity}
\label{appen:constraint}
Constraint ambiguity arises when a query is over-specified (e.g., exact dates, versions, quoted spans) so that retrieval narrows the user’s true intent. We use three complementary signals. First, \textit{total\_hits} \(H(q)\) flags abnormally small result sets, indicating a narrowed scope. Second, the \textit{KL divergence} \(D_{\mathrm{KL}}(P_{\text{top}}\!\parallel P_{\text{corpus}})\) measures how skewed the top-snippet word distribution is relative to the background corpus, revealing over-reliance on special tokens (dates, numbers, quoted phrases). Third, the \textit{relax\_delta\_ratio} \(\rho(q)=\tfrac{H(\mathrm{relax}(q))}{H(q)}\) is an intervention-style cue: it asks how much the hit count jumps when we remove exactly one constraint (a date, a number, or a quoted span). In combination, \(\,H(q)\) low, \(D_{\mathrm{KL}}\) high, and \(\rho(q)\) high strongly suggest over-specialization–induced recall failure, whereas low \(H(q)\) with low \(\rho(q)\) points to genuinely sparse topics rather than over-specification. These cues reduce false positives and guide the LLM toward expert, evidence-aware judgments.

\section{Prompt Templates}
\label{appen:prompt}
This section summarizes the prompt templates used to construct MARCH and CLARION. For each ambiguity type, we provide templates for detection, clarification, answer generation (short/long), and query decomposition from \Cref{fig:prompt-syn-detect,fig:prompt-syn-clar,fig:prompt-general-detect,fig:prompt-general-clar,fig:prompt-sem-detect,fig:prompt-sem-clar,fig:prompt-short-answer,fig:prompt-long-answer,fig:prompt-decompose}.

\clearpage
\newtcolorbox{PromptBox}[1][]{%
  enhanced, breakable,
  width=0.92\linewidth,
  colback=black!12, colframe=black!60,
  boxrule=0.8pt, arc=2pt,
  left=1em, right=1em, top=0.7em, bottom=0.7em,
  before upper={\par\sloppy},
  #1}

\newcommand{\B}[1]{\textbf{#1}\\[0.9\baselineskip]}

\newenvironment{PromptFigure}[2][]{%
  \begin{figure*}[t]
  \centering
  \def\PromptTmpLabel{#1}%
  \def\PromptTmpCaption{#2}%
  \begin{PromptBox}
}{%
  \end{PromptBox}
  \caption{\PromptTmpCaption}%
  \if\relax\detokenize{\PromptTmpLabel}\relax\else
    \label{\PromptTmpLabel}%
  \fi
  \end{figure*}
}

\begin{PromptFigure}[fig:prompt-syn-detect]{Prompt template for syntactic ambiguity detection.}
You are a linguistics expert. \\

1) Read the sentence below.\\
2) Decide whether it is syntactically ambiguous under any of the 18 phenomena.\\
3) If ambiguous, list all applicable phenomenon numbers (ascending). \\

\B{Phenomena (1–18)}
1. PP Attachment (including instrument vs. attribute "with"); 2. Relative-Clause Attachment; 3. Coordination Scope (and/or); 4. Comparative Attachment / Ellipsis; 5. Quantifier / Negation Scope; 6. Dangling / Misplaced Modifier; 7. Genitive-Chain Attachment; 8. Complement vs. Adjunct; 9. Gerund vs. Participle; 10. Ellipsis / Gapping; 11. If-clause Attachment; 12. Right-Node Raising; 13. Adjective Stacking / Coordination; 14. Inclusive vs. Exclusive "or"; 15. Adverbial Attachment (VP vs. S); 16. Focus / Only-scope; 17. Apposition vs. Restriction; 18. Degree / Comparative subdeletion. \\

\textbf{Question}: QUESTION \\

\textbf{Output (JSON)}: 
{ "is\_ambiguous": "Y", "categories": [1, 3, 7] }   // [] if "N" \\

Keys must be exactly "is\_ambiguous" and "categories". No extra text.
\end{PromptFigure}

\begin{PromptFigure}[fig:prompt-syn-clar]{Prompt template for syntactic clarification.}
You are a linguistics expert. \\

The question below is syntactically ambiguous. Write at least {{MIN\_VERSIONS}} distinct clarified questions, each encoding a different structural reading (attachment, scope, etc.). Preserve the topic; each rewrite must be fully unambiguous; concise, natural English. \\

\textbf{Question}: QUESTION \\

\textbf{Output (JSON)}: 
{ "clarified\_queries": ["...", "..."] } \\

Key exactly "clarified\_queries"; provide at least 2 strings; no other keys.
\end{PromptFigure}

\begin{PromptFigure}[fig:prompt-general-detect]{Prompt template for constraint (over-specific) ambiguity detection.}
You are a linguistics expert. \\

1) Read the search query and three RAW metric values.\\
2) Decide if the query shows constraint ambiguity (over-specific constraints harming recall).\\
3) Output ONLY the JSON object in the required format.\\

A query with constraint ambiguity (over-specific) is narrowly constrained
(dates, version numbers, quoted strings, etc.), likely missing the broader intent. \\

\B{Metrics}
\textbf{Total\_hits}: Result count for the literal query.\\
\textbf{KL\_divergence}: D\_KL between top-k snippet unigrams and the whole corpus.\\
\textbf{Relax\_delta\_ratio}: Largest fold-increase in hits after removing one numeric/date/quoted constraint. \\

\textbf{Question}: QUESTION \\

\B{Raw metric values}
\textbf{Total\_hits}: {{TOTAL\_HITS}}\\
\textbf{KL\_divergence}: {{KL\_DIVERGENCE}}\\
\textbf{Relax\_delta\_ratio}: {{RELAX\_DELTA\_RATIO}} \\

\textbf{Output (JSON)}: 
{ "is\_ambiguous": "Y" }   // "N" if not constraint \\

Use expertise; no hard thresholds. No markdown, code fences, or extra keys.
\end{PromptFigure}

\begin{PromptFigure}[fig:prompt-general-clar]{Prompt template for general clarification.}
You are an information-retrieval and linguistics expert. \\

Rewrite the query below into at least {{MIN\_VERSIONS}} broader, faithful variants that
surface the user’s core intent and remove needless specificity or indirections. \\

\B{How to clarify}
1) Identify the core question (fact or relationship truly sought).\\
2) Resolve or drop cascading indirections (replace "the country where X was born" with the direct entity if obvious; else use a neutral placeholder).\\
3) Remove or soften excessive constraints (exact dates, versions, quoted titles).\\
4) Keep the answer type the same; do not over-broaden. Write concise English. \\

\textbf{Question}: QUESTION \\

\textbf{Output (JSON)}: 
{ "clarified\_queries": ["...", "..."] }\\

Key must be exactly "clarified\_queries"; provide at least 2 strings; no extra keys.
\end{PromptFigure}

\begin{PromptFigure}[fig:prompt-sem-detect]{Prompt template for semantic ambiguity detection.}
You are a linguistics expert. \\

Semantically ambiguous lacks sufficient context so multiple reasonable meanings or referents are possible (unclear pronoun, vague time, polysemy, etc.). \\

1) Read the sentence.\\
2) Output "Y" if semantically ambiguous, else "N".\\

\textbf{Question}: QUESTION \\

\textbf{Output (JSON)}: 
{ "is\_ambiguous": "Y" }   // "N" if unambiguous \\

Key must be exactly "is\_ambiguous". No extra text.
\end{PromptFigure}

\begin{PromptFigure}[fig:prompt-sem-clar]{Prompt template for semantic clarification.}
You are a linguistics expert. \\

Rewrite the semantically ambiguous question into at least 2 distinct clarified questions, each resolving a different interpretation. Preserve the original topic. Add only minimal context (time, referent, sense) to make each unambiguous. Concise English. \\

\textbf{Question}: QUESTION \\

\textbf{Output (JSON)}: 
{ "clarified\_queries": ["...", "..."] }\\

Key exactly at least 2 "clarified\_queries"; no other keys.
\end{PromptFigure}

\begin{PromptFigure}[fig:prompt-short-answer]{Prompt template for short answer generation (extractive).}
You are an extractive QA assistant. \\

Given a question and one passage, return the shortest exact span in the passage that answers the question. If no answer, return an empty string. \\

\textbf{Question}: QUESTION \\

\textbf{Passage}: PASSAGE \\

\textbf{Output (JSON)}: 
{ "short\_answer": "..." }\\

Extractive only (verbatim span); no justification or extra text.
\end{PromptFigure}

\begin{PromptFigure}[fig:prompt-long-answer]{Prompt template for long answer generation (merge A1 + A2).}
You are an expert open-domain QA assistant. \\

Combine two validated short answers (A1, A2) to create a single, coherent long answer to the original ambiguous question (OQ). If both can be true, merge into 1–3 fluent sentences. Do not invent facts beyond A1 and A2. \\

Return only JSON that matches the schema: {{SCHEMA}} \\

\textbf{Question}: QUESTION \\

\textbf{Clarified Q1 | Short Answer A1}\\[0.6\baselineskip]
{{CQ1}}\\
A1 = {{A1}} \\

\textbf{Clarified Q2 | Short Answer A2}\\[0.6\baselineskip]
{{CQ2}}\\
A2 = {{A2}} \\

\textbf{Output (JSON)}: 
{ "long\_answer": "..." }\\
\end{PromptFigure}

\begin{PromptFigure}[fig:prompt-decompose]{Prompt template for query decomposition (ordered single-hop bullets).}
You are an information-retrieval expert. \\

Break the complex question into the minimal set of atomic, single-hop sub-questions in the exact order needed to fully answer it. \\

- Output each sub-question as a Markdown bullet starting with "* ".\\
- Each sub-question must ask for exactly one fact or relationship.\\
- No explanations or extra text.\\

\textbf{Question}: QUESTION \\

\textbf{Output (JSON)}: 
{ "sub\_query": "..." }\\
\end{PromptFigure}

\begin{PromptFigure}[fig:prompt-ambiguity]{Prompt template for ambiguity detection and typing (strict JSON).}
You are an expert at analyzing query ambiguity. \\
Your task is to determine whether a query is ambiguous and to classify the ambiguity type. \\

Analyze the following query and decide: \\

1. Provide brief reasoning. \\
2. Is the query ambiguous? \\
3. Which specific aspects make it ambiguous? \\
4. What extra information would clarify it? \\
5. Classify the ambiguity as one of: \textit{"syntactic"}, \textit{"constraint"}, \textit{"semantic"}, or \textit{"none"}. \\

\textbf{Definitions}: \\

* \textbf{syntactic}: multiple plausible grammatical parses (attachment/scope/coordination/pronoun reference). \\
* \textbf{constraint}: over-specific query where a broader, closely related formulation better matches the user's need. \\
* \textbf{semantic}: syntax is clear but meaning/intent admits multiple valid interpretations via world knowledge. \\

\textbf{Query}: \{query\} \\

\textbf{Return STRICT JSON}: \\
\{ \\
\ \ \ "reasoning": "string", \\
\ \ \ "is\_ambiguous": true/false, \\
\ \ \ "ambiguity\_type": "syntactic" \textbar\ "constraint" \textbar\ "semantic" \textbar\ "none", \\
\ \ \ "ambiguous\_aspects": ["..."], \\
\ \ \ "clarification\_needed": "string" \\
\}
\end{PromptFigure}

\begin{PromptFigure}[fig:prompt-clarify]{Prompt template for generating two clarified queries from an ambiguity analysis.}
You are an expert at clarifying ambiguous queries. \\
Given the original query and an ambiguity analysis, rewrite the query into \textbf{two} specific, actionable, and faithful clarified versions. \\

\textbf{Original Query}: \{query\} \\
\textbf{Ambiguity Analysis (JSON)}: \{analysis\} \\

\textbf{Write STRICT JSON}: \\
\{ \\
\ \ \ "reasoning": "why these clarifications resolve the ambiguity", \\
\ \ \ "clarified\_query1": "string", \\
\ \ \ "clarified\_query2": "string" \\
\}
\end{PromptFigure}

\begin{PromptFigure}[fig:prompt-react]{Prompt template for ReAct-style retrieval and answering with a bounded search budget.}
You are a research assistant following ReAct (Reasoning, Acting, Observing). \\

\textbf{Available Actions}: \\

* \texttt{SEARCH[query]} $\rightarrow$ run a search using the configured method \\
* \texttt{ANSWER[planning]} $\rightarrow$ run a planning agent \\
* \texttt{ANSWER[text]} $\rightarrow$ provide a final answer now \\

\textbf{Constraints}: \\

* Max searches allowed: {max\_searches} \\
* Searches used so far: {current\_searches} \\
* Do \textbf{not} reuse the exact same search query as previously used in context. \\

\textbf{Task Query}: \{query\} \\
\textbf{Previous Context}: \\
\{context\} \\

\textbf{Instructions}: \\

1. THINK about the next best step. \\
2. If more evidence is needed, choose \texttt{SEARCH[very specific query]}. \\
3. If sufficient, choose \texttt{ANSWER[concise, well-supported answer]}. \\
4. If you have already reached the maximum allowed searches, you \textbf{must} output \texttt{ANSWER[...]} now. \\

Respond in \textbf{EXACT} format: \\
\texttt{THOUGHT: <your internal reasoning, one short paragraph>} \\
\texttt{ACTION: SEARCH[...specific query...]} \textbf{OR} \\
\texttt{ACTION: PLANNING[...call planning agent...] } \textbf{OR} \\
\texttt{ACTION: ANSWER[...final answer...] } \\
\end{PromptFigure}

\end{document}